\documentclass[times]{elsarticle}
\usepackage{lmodern} % \documentclass[5p,times]{elsarticle}

\usepackage[utf8]{inputenc}
\usepackage[T1]{fontenc}

\let\keptmaketitle\maketitle
\usepackage[colorlinks=true,linkcolor=black,urlcolor=blue,citecolor=blue,anchorcolor=black]{hyperref}
\let\maketitle\keptmaketitle

\usepackage{mathnotation}
\usepackage{amsmath,amssymb,amsthm,amsfonts}
\usepackage[inline]{enumitem}
\usepackage{color}															 % für Farben im \usepackage[table]{xcolor}
\usepackage{xspace}
\usepackage{graphicx}

\allowdisplaybreaks

\usepackage{multirow}
\usepackage[table,xcdraw]{xcolor}
\usepackage[font=small,labelfont=bf]{caption}
\newenvironment{myquote}%
{\list{}{\leftmargin=0.3in\rightmargin=0.3in}\item[]}%
{\endlist}

\journal{ArXiv}
\usepackage[draft]{fixme}
\usepackage[%disable,
backgroundcolor=white,textsize=footnotesize]{todonotes}

\newtheorem{definition}{Definition}
\newcommand{\protege}{Prot\'{e}g\'{e}\xspace}
\newcommand{\stt}[1]{\texttt{\small #1}}

\newcommand{\tax}{\mathit{ax}\xspace}

\newcommand{\mo}{\mathcal{O}\xspace}
\newcommand{\mot}{\mathcal{O^*}\xspace}

\newcommand{\md}{\mathcal{D}\xspace}
\newcommand{\mb}{\mathcal{B}\xspace}
\newcommand{\Te}{P}
\newcommand{\Tne}{N}
\newcommand{\te}{p}
\newcommand{\tne}{n}
\newcommand{\qry}{Q}

\newcommand{\mD}{{\bf{D}}\xspace}

\begin{document}

\begin{frontmatter}

%\title{On the Usefulness and Challenges of Query-based Knowledge Base Debugging Methods}
\title{Are Query-Based Ontology Debuggers Really Helping Knowledge Engineers?\tnoteref{mytitlenote}}
\tnotetext[mytitlenote]{This is a preprint of the work \protect\cite{rodler2019userstudy} that is formally published in the \emph{Knowledge-Based Systems} journal.}
%\tnoteref{t1}

%\author{Patrick Rodler, Dietmar Jannach, Konstantin Schekotihin, and Philipp Fleiss}
\author[]{Patrick Rodler\corref{cor1}}
\author[]{Dietmar Jannach}
\author[]{Konstantin Schekotihin}
\author[]{Philipp Fleiss}
\address{Department of Applied Informatics \\University of Klagenfurt, 9020 Klagenfurt, Austria \\ {[firstname.lastname]}@aau.at}

\cortext[cor1]{Corresponding author}
%\tnotetext[t1]{This work was supported by the Carinthian Science Fund (KWF), contract KWF-3520/26767/38701.}

% OLD abstract start %%%%%%%%%%%%%%%%%%%%%%%%
%\begin{abstract}
%With the increasing spread of Semantic Web technology, applications that are built upon explicitly encoded domain knowledge have regained a certain level of popularity in recent years.
%Since the underlying knowledge bases can easily become large and complex, it is not uncommon that such knowledge bases contain faults.
%Correspondingly, a number of knowledge base debugging approaches, in particular for ontology-based systems, were proposed throughout the last years.
%Query-based debugging is an interactive approach that suggests knowledge engineers to answer a series of questions and then uses the provided answers to localize the true cause of an observed problem.
%Since typical simulation-based evaluations cannot fully inform us about the usefulness of such approaches, we conducted different user studies to assess the practical value and the limitations of such an interactive approach.
%One main insight from the studies is that query-based debugging is indeed more efficient than an alternative algorithmic debugging approach based on test cases.
%We however also observed that users frequently made errors in the process, which highlights the importance of a careful design of the queries that are asked to the users.
%
%\end{abstract}
% OLD abstract end %%%%%%%%%%%%%%%%%%%%%%%%
\begin{abstract}
Real-world semantic or knowledge-based systems, e.g., in the biomedical domain, can become large and complex. Tool support for the localization and repair of faults within knowledge bases of such systems can therefore be essential for their practical success. Correspondingly, a number of knowledge base debugging approaches, in particular for ontology-based systems, were proposed throughout recent years.
Query-based debugging is a comparably recent interactive approach that localizes the true cause of an observed problem by asking knowledge engineers a series of questions. Concrete implementations of this approach exist, such as the \emph{OntoDebug} plug-in for the ontology editor \protege.

To validate that a newly proposed method is favorable over an existing one, researchers often rely on simulation-based comparisons. Such an evaluation approach however has certain limitations and often cannot fully inform us about a method's true usefulness.
% of a newly proposed method.
%However, simulation-based evaluations of such implementations reported in the literature have their limitations. Specifically, they cannot fully inform us about the increased usefulness of a newly proposed method when compared to an existing one.
%if one method is truly more helpful for users during debugging than another one.
%to directly assess the usability improvements the debugging methods.
We therefore conducted different user studies to assess the practical value of query-based ontology debugging.
One main insight from the studies is that the considered interactive approach is indeed more efficient than an alternative algorithmic debugging based on test cases.
We also observed that users frequently made errors in the process, which highlights the importance of a careful design of the queries that users need to answer.
\end{abstract}


\end{frontmatter}

\section{Introduction}
Systems that are built upon Artificial Intelligence (AI) techniques are often classified into two categories:
\begin{enumerate*}[label=\textit{(\roman*)}]
	\item systems that automatically \emph{learn from data} and
	\item systems based on explicitly \emph{encoded domain knowledge} and automated inference services.
\end{enumerate*}
Knowledge-based software systems are typical representatives of the latter form of AI with a number of successful applications in various domains such as planning and scheduling,
%ontology-based systems and Semantic Web applications,
medical advice-giving systems, product configuration, or recommender systems \cite{ModernApproach,DLHandbook,pinedo2016scheduling,felfernig2014knowledge,Jannach2010}.
% DJ: say something about their relevance? despite the recent ML hype ..

The correctness of the decisions and suggestions made by a knowledge-based system depends directly on the ability of an expert to formulate and maintain a knowledge base (KB) that describes the application domain.
Both knowledge formalization and maintenance can be challenging due to
\begin{enumerate*}[label=\textit{(\roman*)}]
	\item the cognitive complexity of the task and
	% KS do not know any good replacement for ``complexity'' maybe ``expressivity'', but it sounds not good.
	\item the size and complexity of the resulting knowledge base---e.g., biomedical ontologies as found on BioPortal\footnote{\url{http://bioportal.bioontology.org}} sometimes contain thousands of axioms.
\end{enumerate*}
% KS leave ``make faults'' (mistakes is not good)
The results reported, e.g., in \cite{Ceraso71,Johnson1999,Rector2004,Roussey2009} suggest that people often make mistakes %systematic faults
when writing or interpreting logical sentences.
Furthermore, in some cases, knowledge bases are constructed in a collaborative manner by multiple contributors, which is another potential source of faults~\cite{Noy2006a,ji2009radon,meilicke2011}.

Overall, given that unintended or contradictory specifications are likely to occur in such knowledge bases, it is essential to provide experts with appropriate tools for fault detection, localization, and repair.
Over the last decades researchers suggested different techniques and implemented a number of assistive tools for these tasks.
Many of these techniques are based on the principles of model-based diagnosis (MBD) \cite{Reiter87}, which is a versatile fault localization method with a range of applications, e.g., in the context of electronic circuits, declarative programs, knowledge bases and ontologies, workflow specifications, as well as programs written in domain-specific and general-purpose languages ~\cite{paper:felfernig:2004,DBLP:conf/aadebug/MateisSWW00,DBLP:conf/semweb/FriedrichS05,JannachSchmitz2014,Friedrich1999,10.1007/BFb0019402,Rodler2015phd,dekleer1987}.

In the context of knowledge base debugging, MBD techniques are applied when a knowledge base does not fulfill some basic requirements, e.g., when it is inconsistent in itself or when test cases indicate a failure, i.e., an unexpected output.
In the usual MBD problem formulation, test cases are logical sentences that the
intended
%correctly formulated
knowledge base must (or must not) entail. %under the assumption of a .
The output of an MBD tool is a collection of diagnoses, where each \emph{diagnosis} corresponds to a set of assumedly faulty parts of the knowledge base. Users of the debugger, such as experts or knowledge engineers, can then investigate one diagnosis after another and inspect the involved components to see if they are faulty or not.

Unfortunately, the number of diagnoses can in some cases be large, e.g., because the information provided by the test cases is insufficient and does not allow the debugger to isolate the true cause of the observed failure.
In such cases, already early works suggested asking an expert to provide additional information to narrow down the set of possible fault locations.
% KS Moved the citation to the next sentence.
For example, in the traditional application domain of MBD techniques---electronic circuits---users of a diagnosis system are asked to make additional measurements that give some indication of the health state of certain components~\cite{dekleer1987}.
In more recent years, different algorithms for \emph{sequential} (or: interactive) diagnosis of knowledge-based systems were proposed~\cite{meilicke2011,DBLP:conf/ecai/ShchekotykhinFRF14,Rodler2011,Shchekotykhin2012}.
Debuggers of this type interactively ask users to provide feedback about the correctness of parts of the knowledge base or certain inferences.
One concrete implementation of such a debugger is \emph{OntoDebug}
%\footnote{\url{http://isbi.aau.at/ontodebug/}} 
\cite{DBLP:conf/foiks/SchekotihinRS18,DBLP:conf/icbo/SchekotihinRSHT18a}, a plug-in for the \emph{\protege} ontology editor \cite{noy2003protege}.
%\footnote{\url{https://protege.stanford.edu/}}
% KS leave ``true cause''
Compared to approaches that solely rely on test cases, the main advantage of such \emph{query-based} techniques is that they can interactively guide their users to the true cause of the observed problem.
% KS comments are irrelevant (diangosis cardinality >1 is supported)
In addition, if users always provide correct answers to the debugger's questions, then query-based diagnosis techniques can guarantee the identification of the true fault location.

The evaluation of sequential diagnosis techniques is usually based on simulations designed to measure, for instance, the time needed to derive the best next query to the expert or the total number of required queries to isolate a fault.
Such measures can however have certain limitations when assessing the true usefulness of a debugging approach.
In the domain of software engineering, the practical relevance of results obtained with the help of simulation-based evaluations of debugging tools was previously questioned by Parnin and Orso in \cite{Parnin:2011:ADT:2001420.2001445}.
In recent years, a number of user studies were therefore conducted
%by researchers
that directly assess the usefulness of different academic approaches to tool-supported testing and debugging in the context of software engineering \cite{DBLP:conf/euromicro/RamlerWS12,DBLP:conf/issta/StaatsHKR12,DBLP:conf/issta/FraserSMAP13}.

With the present paper, we continue this line of research.
Specifically, our goal was to assess the usefulness of query-based approaches for knowledge base debugging.
We correspondingly conducted \emph{laboratory} studies in the form of testing and debugging exercises that were specifically designed to evaluate if query-based debugging is truly favorable over a previous debugging approach based on test cases.
%the interactive method in comparison to the approach based on test cases.
%For each exercise participants either applied an approach based on test cases or were supported by a query-based debugger, concretely, by an earlier version of the OntoDebug tool mentioned above.
Our corresponding research questions are therefore related to
\begin{enumerate*}[label=\textit{(\roman*)}]
	\item the efficiency and effectiveness of query-based debugging (i.e., do experts need less time, do they find more faults?),
	% KS effectiveness is not good here
	\item the cognitive ability of users to find out which of the returned diagnoses is the correct one, and
	% we mention in Section 1 that users are experts or knowledge eng.
	\item the difficulty of answering system-generated queries for experts.
\end{enumerate*}

Among other aspects, our results
%will
indicate that a query-based approach
%is indeed helpful to
can make the debugging process more efficient, without leading to a loss in effectiveness.
Furthermore, our experiments and previous studies show that experts sometimes provide wrong answers to the questions of a debugger (``oracle errors''). We therefore conducted additional pen-and-paper studies to develop and validate a prediction model that can be used to estimate the probability of oracle errors based on the cognitive complexity of a query or a test case.

The paper is organized as follows. After discussing previous works in Section \ref{sec:related-work}, we provide the technical background on MBD-based knowledge base debugging in Section \ref{sec:kb-debugging}. Section \ref{sec:research-questions} presents the detailed research questions of our work, and Section \ref{sec:study1} and Section \ref{sec:study2} discuss the outcomes of our main studies. In Section \ref{sec:prediction-model}, we finally present first results regarding our prediction model for oracle errors. The paper ends with a discussion of research limitations and a summary of our contributions. % and an outlook on future works.
% \todoil{Check text at the end.}

\section{Related Work}
\label{sec:related-work}
The process of creating and maintaining a KB is prone to error and---like in standard software development projects---experts can make mistakes when they encode the knowledge about a problem domain. Correspondingly, a number of techniques and tools for KB testing and debugging were proposed over the years. In the following, we first briefly review the main debugging strategies suggested in the literature and then specifically discuss previous works that aim at evaluating the utility of the corresponding \emph{tools} with the help of user studies.

\subsection{General Knowledge Base Debugging Approaches}
We can mainly distinguish between \emph{model-based} and \emph{heuristic} approaches for KB debugging. Among the \emph{model-based} approaches, those based on the general MBD principles proposed in \cite{Reiter87} are probably the most popular ones. They have, for example, been used to debug ontologies \cite{DBLP:conf/semweb/FriedrichS05,Kalyanpur.Just.ISWC07,Horridge2008}, constraints \cite{paper:felfernig:2004,Junker04}, or Answer Set Programming encodings \cite{DBLP:conf/wlp/GebserPSTW07,DBLP:journals/tplp/OetschPT10}.

In case of ontology debugging, MBD methods are used to find sets of axioms, called diagnoses (or: candidates/repairs), that must be modified by a developer in order to formulate the intended ontology.
From the technical perspective these methods can roughly be classified in glass-box and black-box ones~\cite{ParsiaSK05,SchlobachHCH07}.
Glass-box approaches~\cite{SchlobachC03,Kalyanpur2006a,BaaderP07, BaaderP10,ChengQ11,DBLP:conf/sum/OzakiP18,DBLP:conf/cade/KazakovS18} modify the reasoner such that a single execution run outputs justifications or diagnoses directly.
%\todoil{''run'' might be misinterpreted as the run of a debugger, and not of a reasoner. Maybe write ``reasoner run'' (although it repeats the word ``reasoner'') ??}
Black-box methods, in contrast, usually apply various search techniques~\cite{DBLP:conf/semweb/FriedrichS05,DBLP:conf/ecai/ShchekotykhinFRF14,SchlobachHCH07,DBLP:journals/ai/PenalozaS17} with calls to highly-optimized reasoners for consistency checking and/or the computation of irreducible faulty subsets of an ontology, called justifications or conflicts  \cite{Horridge2008,Junker04,KalyanpurPHS07,DBLP:conf/aaai/Shchekotykhin15}.
%\todoil{justifications appear (when explaining glass-box) before they are explained -- maybe exchange the order of expplaining black-box and glass-box to resolve?}

In practical settings, given an inconsistent/incoherent ontology, an MBD approach might return more than one diagnosis. % (sometimes referred to as \emph{candidates}).
In order to restrict the number of obtained diagnoses to only relevant ones, Friedrich et al.\ \cite{DBLP:conf/semweb/FriedrichS05} suggested the notion of MBD test cases, which were later also used in, e.g., \cite{DBLP:journals/ws/ShchekotykhinFFR12,DBLP:conf/kr/GrauJKZ12,DBLP:conf/tableaux/FurbachS13}.
Each test case is defined as a (set of) axiom(s) that the intended ontology \emph{must} or \emph{must not} entail. %ed by the intended ontology.
%Given a set of test cases a debugger can use them to focus only on those diagnoses such that if all axioms of a diagnosis are changed the resulting ontology will satisfy all test cases.
A debugger can then use these test cases to focus only on those diagnoses for which it is guaranteed that a (suitable) modification of all the axioms of a diagnosis will result in an ontology that satisfies all test cases.
However, in many situations, it can be unclear to a developer which test cases should be formulated before the diagnosis session such that a debugger will be able to find the true cause of an unexpected output.
In this case, query-based approaches \cite{dekleer1987,rodler_jair-2017,Shchekotykhin2012,rodler17dx_activelearning} help the user to automatically create test cases. Specifically, the task of the users is reduced to answering a sequence of queries on whether or not the intended ontology must entail a given set of axioms.
Assuming that all answers of the developer are correct, a sequential debugger can determine the \emph{true diagnosis} within the candidates, i.e., the one diagnosis that pinpoints the actually faulty parts of the knowledge base.

Depending on the complexity of the underlying problem, model-based methods can be comparably costly in terms of computation time and space.
However, one main advantage of MBD %-based
approaches is that any diagnosis which is returned is a precise and succinct explanation of all identified problems. % (bugs).

In contrast, \emph{heuristic approaches} to KB debugging, such as \cite{WangHRDS05,DBLP:conf/f-egc/RousseyCSSB12}, are usually based on handcrafted syntactic pattern matching procedures, see, e.g., \cite{Rector2004,DBLP:journals/jamia/RectorBS11}.
Their main advantage is that they allow for fast fault localization in case model-based approaches are too slow.
Typically, these debugging procedures are designed to find (combinations of) syntax constructs in a KB that are highly likely to be faulty.
Examples of such constructs are, among others, the application of universal role restrictions and disjointness constraints in related ontology axioms \cite{Roussey2009}.
%\todoil{''...AND disjointness constraints...'', right? I.e., the conveyed meaning should be that related axioms become problematic (as per this pattern) if BOTH univ role restr and disj constr are used, right?}
Although these methods are computationally efficient, they are often \emph{incomplete} (i.e.,
they can only identify bugs for which appropriate heuristics were defined) and sometimes \emph{unsound} (i.e., they might return diagnoses that comprise actually correct axioms).
%Consequently, the reliability of the returned results in terms of precision and recall
%, e.g., if the found conjectured-wrong logical sentence is really a bug,
%can be low.
%TODO should we mention ``precision and recall'' here? or leave the sentence as ``...the reliability of the returned results can be low.'' ?

In this paper, we focus on the MBD approach presented in \cite{Rodler2015phd,Shchekotykhin2012}, since it
\begin{enumerate*}[label=\textit{(\roman*)}]
	\item provides guarantees about the completeness and soundness of the debugging algorithms and
	\item allows for a precise fault localization by querying its users for additional information.
\end{enumerate*}

\subsection{Usefulness Analysis of Tools}
Since KBs in practice can be large and complex, the research community developed a number of Integrated Development Environments (IDEs) for KB creation and maintenance.
Examples of such environments are the MiniZinc IDE for constraint modeling \cite{DBLP:conf/cp/NethercoteSBBDT07}, \protege, which supports the creation of ontologies \cite{Musen2015}, ASPIDE as a tool for the development of Answer Set Programs \cite{DBLP:conf/lpnmr/FebbraroRR11}, as well as various Prolog IDEs like SWI-Prolog \cite{DBLP:journals/tplp/WielemakerSTL12}.
Several of these IDEs come with embedded debugging support or can be extended with  external tools like the OntoDebug plug-in used in this paper \cite{dodaro2015interactive,DBLP:conf/cpaior/LeoT17,DBLP:conf/foiks/SchekotihinRS18}.

Two main approaches exist in the literature to evaluate the usefulness of KB debugging tools.
The first one conducts \emph{computational analyses} providing insights about the usefulness of the tools indirectly.
The second form is based on \emph{user studies}, where the performance and behavior of experts while using the debugger is observed and analyzed.
Most of the research in the field is based on the first form of experiments.
In comparison to user studies, conducting computational analyses is usually easier, since the only requirement for such evaluations is the existence of a representative collection of knowledge bases that contain real-world or %artificially
injected faults.
Given such KBs, the performance of different debugging algorithms can be compared, for example, in terms of their time and space complexity, the number of calls to the reasoner, the theoretical number of required user interactions, or the precision of the fault localization process.
The obtained results can then be used to \emph{indirectly} assess if a given debugging approach is favorable over another.
For instance, we can assume that the reduction of the required computation time increases the usefulness of a system, e.g., because the developer gets faster feedback and can find more bugs in a shorter time.

However, such computational analyses have their limitations. They, for example, cannot be used to determine if certain assumptions made by the evaluated debugging methods actually hold. For instance, the interactive ontology debugging method suggested in \cite{Shchekotykhin2012} assumes that a user can decide with certainty if the intended ontology must entail an arbitrary axiom or not. If this assumption does not hold, i.e., the user cannot (correctly) answer all queries of the debugger, the fault localization process might not lead to a unique (correct) result.

User studies can help us to verify such assumptions and can give us additional insights regarding the acceptance and true usefulness of a debugging tool. In the literature, only a few examples of such user studies exist.
For instance, the model-based ontology debugging approach proposed in \cite{Kalyanpur.Just.ISWC07} and implemented in the Swoop editor \cite{DBLP:journals/ws/KalyanpurPSGH06} was evaluated by twelve undergraduate and graduate students \cite{Kalyanpur2006a}.
The authors' goal was to investigate if providing \emph{justifications} for certain inferences can help users find and repair bugs more efficiently.
Every subject that participated in the study had at least nine months of experience in ontology engineering and went through an additional 30-minute training session on ontology debugging.
The results of the study indicate that tool support in the form of justifications during the debugging process is essential for successful fault localization. However, given the small number of participants, the authors were not able to validate that their results are statistically significant.

Another user study reported in \cite{Horridge2011b} investigated if justifications generated by model-based ontology debuggers can actually be understood by users.
%Experiments were conducted with 14 undergraduate students and their results showed that justifications can be separated into easy and hardly understandable.
Experiments were conducted with 14 undergraduate students and their results showed that justifications can be separated into (cognitively) easy and hard ones.
Unfortunately, also in this case the small number of participants did not allow the authors to obtain sufficient statistical evidence to understand why the users find some explanations hard or easy to comprehend.

Finally, a collection of heuristic approaches \cite{Rector2004,Roussey2009,DBLP:conf/ekaw/Svab-ZamazalS08} was studied in \cite{corcho2009pattern} and compared with an MBD approach \cite{Horridge2008}.
All 14 subjects participating in the study were educated software engineers and had some experience with ontologies, but no knowledge about hydrology, which was the domain of the study.
The task of the participants was to debug and repair an ontology without understanding exactly what it was about.
One group of six participants was supported by the MBD approach; the remaining subjects used a heuristic strategy.
The obtained results were not fully conclusive.
Both participant groups needed about the same amount of time, and no clear preference for either of the approaches was observed.
Only for the problem of repairing the ontology, the heuristic patterns helped the subjects to identify bugs more accurately.
However, this result must be interpreted with care because the model-based tool did not provide any repair support at that time.

In our work, we continue this line of research which aims to assess the usefulness of debugging approaches based on user studies. Similarly to previous work, we base our user studies on different KBs (ontologies) in which we injected a number of faults.
In addition, like in previous research, we involve students in the studies, who have a certain level of education in the development and debugging of ontologies and who received some initial training with the tool.
In contrast to previous studies, we were able to recruit a larger number of participants, which allows us to apply certain statistical analyses.
Moreover, we are focusing not on justifications, which are alternative explanations of \emph{one} fault, but on diagnoses, where each diagnosis provides a potential characterization of \emph{all} faults in an ontology.

\section{Background: Knowledge Base Debugging with MBD}
\label{sec:kb-debugging}

In this section, we outline the main principles of applying model-based diagnosis techniques for knowledge base debugging. We use the particular problem of ontology debugging to illustrate the problem. Ontology debugging was also the task in the user studies reported in this paper, where the participants used the \emph{OntoDebug}\footnote{\url{http://isbi.aau.at/ontodebug/}}
debugging plug-in \cite{DBLP:conf/foiks/SchekotihinRS18} of the popular ontology editing tool \emph{\protege} \cite{noy2003protege}.\footnote{\url{https://protege.stanford.edu/}}
The underlying principles and algorithms of the debugging approach are, however, not limited to ontologies and can be applied for various forms of knowledge representation and reasoning, see \cite{Rodler2015phd,SchekotihinSchmitzEtAl2016,dodaro2015interactive,rodler17dx_reducing}.

\subsection{Model-based Diagnosis for Ontology Debugging}
%In the field of computer science, ontologies are the core of semantic systems (e.g., within Semantic Web applications) and formally describe, e.g., using a language like OWL \cite{OWL2specification}, the relevant concepts in a domain, their properties and how the concepts are interrelated.
In the field of computer science, ontologies are the core of semantic systems.
%(e.g., within Semantic Web applications) and
Using a language like OWL \cite{OWL2specification}, they formally describe the relevant concepts in a domain as well as their properties and interrelations.
Usually the main goal of semantic applications is to use some form of logic-based reasoning to derive additional facts (\emph{entailments}) from the given knowledge base.

The starting point for a debugging session is normally when we observe a discrepancy between what we call the \emph{intended ontology} (denoted as $\mot$) and
% what we observe for a
the current version of an ontology $\mo$.
Such a discrepancy could be the inconsistency of $\mo$, the unsatisfiability of its classes, or the presence or absence of certain entailments \cite{DBLP:conf/foiks/SchekotihinRS18}.
In the biology domain, a knowledge engineer might, for example, expect that the ontology-based system is able to deduce from the given axioms that men are animals.\footnote{See, e.g., \url{http://owl.man.ac.uk/2003/why/latest/}.}
If, however, it is inferred, e.g., that men and animals are disjoint, the underlying KB is incorrect and the problem is to find one or more faults in the ontological axioms.

\subsubsection{Formal Characterization: Diagnosis Problem}
The automated fault localization process starts with the generation of a diagnosis problem instance, which is formally defined as follows \cite{paper:felfernig:2004,DBLP:conf/foiks/SchekotihinRS18}.

\begin{definition}[Diagnosis Problem Instance (DPI)]\label{def:inst}
	Let $\mo$ be an ontology (a set of possibly faulty axioms) and $\mb$ be a background theory (a set of correct axioms) where $\mo \cap \mb =\emptyset$, and let $\mot$ denote the (unknown) intended ontology.
	Moreover, let $\Te$ and $\Tne$ be sets of axioms where $\mot \cup \mb$ \emph{entails} each $\te \in \Te$ and \emph{does not entail} any $\tne \in \Tne$.
	Then, the tuple $\tuple{\mo, \mb,\Te,\Tne}$ is called a \emph{diagnosis problem instance (DPI)}.
\end{definition}

% KS: not really happy with the sentence below, because it simply repeats the definition.
%A diagnosis is then a set of axioms that are removed from the ontology, with the particular requirement that the resulting ontology, together with the background knowledge and the positive test cases, is
%\begin{enumerate*}[label=\textit{(\roman*)}]
%	\item consistent and
%	\item does not entail the negative test cases.
%\end{enumerate*}

A diagnosis then is a set of axioms such that the removal of these axioms from the ontology, and the subsequent addition of the background knowledge and the positive test cases, yields a consistent (coherent) ontology that satisfies all test cases.

\begin{definition}[Diagnosis]\label{def:diag}	
	Let  $\tuple{\mo, \mb,\Te,\Tne}$ be a DPI.
	Then, a set of axioms $\md\subseteq\mo$ is a \emph{diagnosis} iff both of the following conditions hold:
	\begin{enumerate}
		\item $(\mo \setminus \md) \cup \Te \cup \mb $ is consistent (coherent, if required)\footnote{An ontology $\mo$ is \emph{coherent} iff there do not exist any unsatisfiable classes in $\mo$. A class $X$ is \emph{unsatisfiable} in an ontology $\mo$ iff, for each interpretation $\mathcal{I}$ of $\mo$ where $\mathcal{I} \models \mo$, it holds that $X^{\mathcal{I}} = \emptyset$. See also \cite[Def.~1 and 2]{qi2007measuring}} %TODO: probably we should explain which one is relevant for us.
		\item $(\mo \setminus \md) \cup \Te \cup \mb \not\models \tne$ for all $\tne \in \Tne$
	\end{enumerate}
	A diagnosis $\md$ is \emph{minimal} iff there is no $\md^\prime \subset \md$ such that $\md^\prime$ is a diagnosis.
\end{definition}

Different diagnosis computation algorithms exist; they can be distinguished
based on whether they generate diagnoses \emph{indirectly}, i.e.,
via the computation of conflict sets,
or \emph{directly}, e.g., via divide-and-conquer techniques or through the prior compilation of the problem to an alternative target representation like SAT
\cite{Reiter87,Rodler2015phd,DBLP:conf/ecai/ShchekotykhinFRF14,SchlobachHCH07,rodler2018socs,RodlerH18_dx,darwiche2001decomposable,jiang2003computation,torasso2006,metodi2014}.
% KS added ranking
In addition, the diagnoses can be \emph{ranked} (ordered) according to various criteria, such as their cardinality, i.e., number of axioms in a diagnosis, or their likelihood \cite{dekleer1987}. Such a ranking can simplify the analysis and comparison of diagnoses by allowing the user to focus on the most important ones.

\subsubsection{Example}
We use the following example to illustrate how MBD techniques can be applied to ontology debugging.
Let our ontology consist of the following \emph{terminological axioms}
%Let us assume that our ontology $\mo$ includes the following \emph{terminological axioms} $\mo = $
% TODO proper centering
%\mbox{\hspace{5pt}}
$\{\tax_1 : A \sqsubseteq B,$
$\tax_2 : B \sqsubseteq C,$
$\tax_3 : C \sqsubseteq D,$
$\tax_4 : D \sqsubseteq R\}$.
They define that $A$ is a subclass of $B$, $B$ a subclass of $C$ etc. In a specific domain, this could, e.g., mean that a \emph{MathStudent} is a subclass of \emph{Student}, which is a subclass of \emph{UnivMember}, etc.
Further, the ontology contains two \emph{assertional axioms} $\{\tax_5 : A(v),\; \tax_6 : A(w)\}$, which specify that $v$ and $w$ are instances of class $A$.
%Furthermore, let us assume that the user defined a background theory,  which contains two \emph{assertional axioms} $\mb = \{A(v), A(w)\}$.
%These axioms specify that $v$ and $w$ are instances of class $A$ that should not be considered as fault candidates in the debugging process.
In a practical application, we could have an assertion like \emph{MathStudent(john)}.
Let us assume that the two assertions are known to be correct, and thus should not be considered as fault candidates in the debugging process. To this end, the knowledge engineer would add these axioms to the \emph{background theory} $\mb$.
% \cite{Rodler2015phd}.
That is, the ontology would be split into a possibly faulty part $\mo := \setof{\tax_1,\dots,\tax_4}$ and a correct part $\mb := \setof{\tax_5,\tax_6}$ in this specific case.
To make sure that the ontology is correct, we assume the user specifies a set of \emph{positive test cases} $\Te=\{B(v)\}$ and a set of \emph{negative test cases} $\Tne=\{R(w)\}$, which means that the intended ontology entails that $v$ is of class $B$ and does \emph{not} entail that $w$ is of class $R$.

%Unfortunately, our ontology $\mo$, together with the correct axioms $\mb$, entails $R(w)$ ($\mo \cup \mb \models R(w)$), since $A(w)$ holds and $A$ transitively is a subclass of $R$.
Unfortunately, the ontology $\mo$, together with the correct axioms $\mb$,
entails $R(w)$, i.e., $\mo \cup \mb \models R(w)$, since $A(w)$ holds and $A$ is transitively a subclass of $R$.
Now, given the specified DPI $\tuple{\mo, \mb,\Te,\Tne}$ as an input, a debugging system will identify the following four minimal diagnoses:
$\md_1:\left[\tax_1\right],\;\md_2:\left[\tax_2\right],\;\md_3:\left[\tax_3\right],\;\text{and } \; \md_4:\left[\tax_4\right]$.

\noindent
The intuitive explanation why we get these diagnoses is that
%Intuitively, in the example,
the removal of any individual axiom in $\mo$ would break the subclass relationship chain, and the undesired entailment $R(w)$ would not be present anymore.

However, based on the positive and negative test cases alone, an MBD algorithm cannot discriminate between the four diagnoses, and we cannot derive the \emph{true} cause of the problem. The user can therefore either inspect all diagnoses manually, or provide more information, e.g., in terms of additional test cases.

% KS D_2 etc. are not diagnoses by definition
Assume that the user specifies an additional negative test case $B(w)$. With $\Tne=\{R(w), B(w)\}$ and $\Te=\{B(v)\}$, a debugger will return $\md_1$ as the only minimal diagnosis. Because, the modifications suggested by the sets of axioms $\md_2,\; \md_3, \text{ and }\md_4$ leave $\tax_1$ untouched, and $\tax_1$ in conjunction with $A(w) \in \mb$ leads to the entailment of $B(w)$, and thus to a violation of the negative test cases.

However, the modified ontology $\mo_1 := \mo\setminus\md_1$ now does not entail the positive test case $B(v)$ anymore. Therefore, $\mo_1$ must be extended somehow.
Since the debugger cannot know how to correctly extend the knowledge base, one strategy is to use the required entailments $\Te$ explicitly as an extension \cite{Rodler2015phd}. Hence, in our example, one would simply add $B(v)$ to $\mo_1$.

\subsection{Sequential Diagnosis}
As the example shows, additional knowledge (in our case, test cases) can help to further focus the debugging process and rule out possible fault candidates. Not all test cases are, however, equally helpful.
%One of the goals of \emph{sequential diagnosis} is therefore to automatically identify ``good'' or optimal  test cases, and to interactively \emph{query} the user (or some other \emph{oracle}) if the axioms of each such test case are entailments or non-entailments of the intended ontology.
One of the goals of \emph{sequential diagnosis} is therefore to automatically identify ``good'' or optimal  test cases, and to interactively ask the user (or some other \emph{oracle}) to classify the generated test cases as either positive (intended entailment) or negative (non-intended entailment). We call such a (set of) test case(s) selected by the system and shown to the user for classification a \emph{query}. Based on the user's answer, the debugger can then update its knowledge in terms of the positive and negative test case sets and repeat the process until only one single diagnosis remains.

\subsubsection{Formal Characterization: Oracle and Queries}
The notions of an oracle and a query can formally be described as follows. An oracle categorizes elements of a set of axioms either as positive or negative test cases by checking if the intended ontology must or must not entail these elements.
%\todoil{PR, please check this definition since its condition has no quantifier and therefore can be interpreted ambiguously. For instance, is $ans$ an oracle if only some $\tax \in \mathbf{Ax}$ fulfill the conditions, if all, if none? It seems that there was a try to make a definition for both partial and full oracles, but in my opinion the result is questionable. I would define a partial $ans$ ($\exists \tax \in \mathbf{Ax}$) and then add that $ans$ is full if $\forall \tax \in \mathbf{Ax}$. Also, do we actually need the partial case?}
\begin{definition}[Oracle]\label{def:oracle}
	Let $\mathbf{Ax}$ be a set of axioms. Furthermore, let $ans:\mathbf{Ax}\to \setof{\Te,\Tne}$ be a function which assigns axioms in $\mathbf{Ax}$ to either the positive or the negative test cases.
	Then, we call $ans$ an \emph{oracle} w.r.t.~the intended ontology $\mot$, iff for any $\tax \in \mathbf{Ax}$ both of the following conditions hold:
	\begin{align*}
	ans(\tax) = \Te \quad &\implies \quad\mot \cup \mb \models \tax
	%	\label{eq:ans_pos}
	\\
	ans(\tax) = \Tne \quad &\implies \quad\mot \cup \mb \not\models \tax
	%	\label{eq:ans_neg}
	\end{align*}
\end{definition}

Note that the function $ans$ can either be total or partial. In the first case, the oracle (user) is a \emph{full domain expert} and able to classify all queried axioms; in the latter case, there might be axioms that the oracle is not able to classify.

Since our goal is to narrow down the set of possible diagnoses, a debugger should propose only queries that guarantee the acquisition of \emph{relevant} information. In other words, each query should eliminate at least one diagnosis, given \emph{any} answer of a full domain expert. Generally, a query consists of one or more axioms and can be characterized as follows.\footnote{Whenever we speak of a ``query'' throughout this work, we mean a query in terms of Definition~\ref{def:query}, which must not be confused, e.g., with the concept of a query in terms of a query language such as OWL-QL \cite{fikes2004owl}. In our scenario, queries are answered \emph{based on the knowledge of an oracle about the intended ontology}, with the aim to locate faults in an ontology. Queries in terms of query languages are answered \emph{based on the knowledge specified in an ontology, knowledge graph, etc.} in order to find answers to questions of relevance.}
%a user is asked queries about the intended ontology to find bugs in the current one, whereas query languages are used to }
\begin{definition}[Query]\label{def:query}
	Let $\tuple{\mo,\mb,\Te,\Tne}$ be a DPI, $\mD$ be a set of diagnoses for this DPI, and $\qry$ be a set of axioms. Moreover, let $\qry_{ans}^\Te := \setof{q \in \qry\mid ans(q)=\Te}$ and $\qry_{ans}^\Tne := \setof{q \in \qry\mid ans(q)=\Tne}$ denote the subsets of $\qry$ assigned to $\Te$ and $\Tne$ by an oracle $ans$.
	
	Then we call $\qry$ a \emph{query} for $\mD$ iff, for any classification $\qry_{ans}^\Te,\qry_{ans}^\Tne$ of the axioms in $\qry$ of a full domain expert oracle $ans$, at least one diagnosis in $\mD$ is no longer a diagnosis for the new DPI $\tuple{\mo,\mb,\Te\cup\qry_{ans}^\Te,\Tne\cup\qry_{ans}^\Tne}$.
\end{definition}
Different strategies were proposed in the literature to determine ``good'' or optimal queries, see, e.g.,~\cite{dekleer1987,DBLP:journals/corr/Rodler16a,rodler_singleton-2019}. Usually, this is accomplished by computing a set of diagnoses and by analyzing the effects of applying the different diagnoses with respect to a potential query. Complementary to this approach, a recent work suggests novel ways of diagnosis computation to reduce the user's time and effort for query answering \cite{rodler2018socs}.
%One additional outcome

In general, a byproduct of the process of determining the queries is a \emph{quality} estimate for each resulting query. Such a quality measure can, for example, be based on the expected information gain after the user has answered the query \cite{dekleer1987}, on reinforcement learning \cite{Rodler2013}, or on
criteria \cite{rodler17dx_activelearning,rodler2018ruleML,RodlerS18_dx} adopted from the field of active learning \cite{settles2012}. Finally, since the generation of queries requires potentially costly calls to an underlying reasoner, approaches exist that aim to minimize the number of these computations \cite{Rodler2015phd,SchekotihinSchmitzEtAl2016,rodler-dx17,rodler17dx_activelearning}.

\subsubsection{Example}

One way to assess the utility of different possible test cases---which in the end correspond to queries to the user---is to analyze the entailments of the ontologies $\mo_i^* := (\mo \setminus \md_i) \cup \Te$ after the application of the different diagnoses $\md_i$.

%In our example from above, the four ontologies
%$\mo_1^*,\dots,\mo_4^*$
%have, among others, the following entailments $B(w)$, $C(w)$, $D(w)$ (see Eq.~\ref{ex:O_i^*} below). These entailments can be obtained, e.g., with the help of the realization service \cite{DLHandbook} of a Description Logic reasoner \cite{Shearer2008,sirin2007pellet} and can serve us as test cases.
%\begin{gather}
%\mo^*_1: \emptyset \quad \mo^*_2: \{B(w)\} \quad \mo^*_3: \{B(w),C(w)\}  \quad \mo^*_4: \{B(w),C(w), D(w)\}
%\label{ex:O_i^*}
%\end{gather}
In our example from above, the four ontologies
$\mo_1^*,\dots,\mo_4^*$
have, among others, the following entailments:
%$B(w)$, $C(w)$, $D(w)$ (see Eq.~\ref{ex:O_i^*} below).
$$
\mo^*_1: \emptyset,\;\,\,\, \mo^*_2: \{B(w)\},\;\,\,\, \mo^*_3: \{B(w),C(w)\},\;\,\,\, \text{and} \;\,\,\, \mo^*_4: \{B(w),C(w), D(w)\}
%\label{ex:O_i^*}
$$

\noindent
These entailments can be obtained, e.g., with the help of the realization service \cite{DLHandbook} of a Description Logic reasoner \cite{Shearer2008,sirin2007pellet} and can serve as test cases.

Let us assume that the user knows that $D(w)$ must be entailed and adds it as a positive test case, i.e., the diagnosis problem instance is now
\begin{equation*}
DPI  = \langle\mo,\{A(v),A(w)\}, \{B(v),D(w)\},\{R(w)\}\rangle
\end{equation*}

Given this additional information, a model-based debugger will return only one single diagnosis, $\md_4=[\tax_4]$. All other diagnoses that existed for the problem instance without the new test case, are no longer minimal diagnoses. Specifically, the deletion of each of the diagnoses $\md_1$, $\md_2$, or $\md_3$ from $\mo$ does not affect $\tax_4$, which is however---due to $D(w) \in \Te$---responsible for the unwanted entailment $R(w)\in N$.

Sequential diagnosis algorithms usually make analyses of this type to determine queries (test cases) that are likely to narrow down the set of remaining diagnoses. At the end, the user only has to categorize such system-generated queries and acts as an \emph{oracle} for the debugger.

\subsection{The OntoDebug Plug-In to Prot\'{e}g\'{e}}
\label{sec:ontodebug_plugin}
The described concepts for sequential and test case based MBD for ontologies were implemented in the OntoDebug plug-in \cite{DBLP:conf/foiks/SchekotihinRS18,DBLP:conf/icbo/SchekotihinRSHT18a} of the widely-used \protege ontology editor.
There are two main situations when the user of the tool---possibly after some maintenance activities---might initiate a debugging session with the OntoDebug plug-in. First, the built-in reasoner of \protege might detect that the given ontology is faulty, e.g.\ inconsistent or incoherent, in itself.\footnote{In contrast to other application areas of model-based diagnosis techniques---such as fault localization in electronic circuits \cite{Reiter87,dekleer1987}---inconsistencies can be present in the context of ontology debugging problems without any initially given test cases (observations).}

% KS we use here ``does'' to
Second, even if the ontology in itself is consistent and coherent, the user might want to ensure that the implemented ontology %does
corresponds to the intended one by specifying one or more test cases. If the test cases lead to the disclosure of unexpected entailments, an inconsistency or an incoherency, it is obvious that something is wrong with the ontology.

\begin{figure*}[t]
	\centering
	\includegraphics[width=0.95\linewidth]{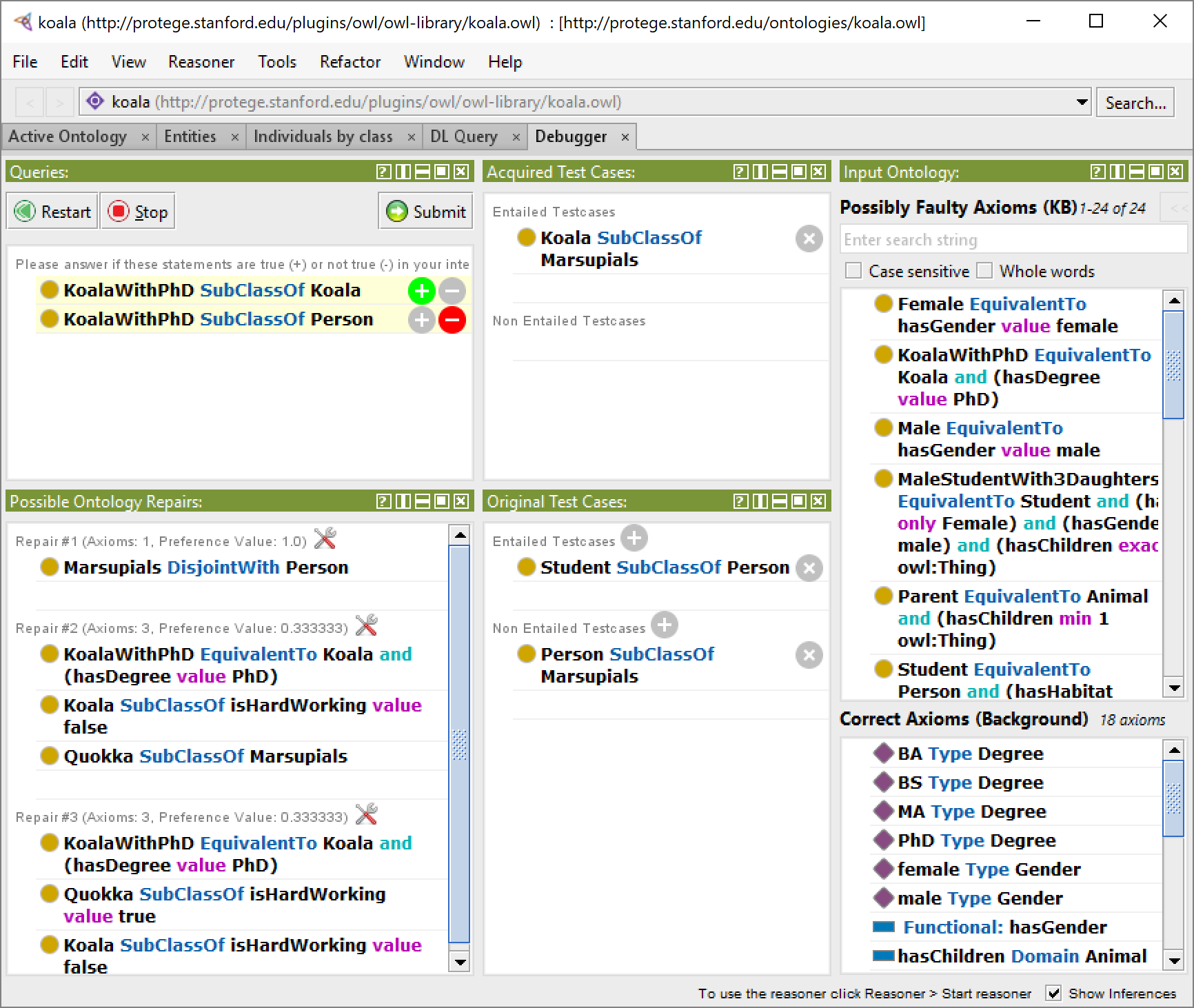}
	\caption{Interactive ontology debugging with the latest version of OntoDebug.}
	\label{fig:koala:session}
\end{figure*}

One possible first step for the user when starting the debugging process with OntoDebug---independent of how the user detected that there is a problem---is to tell the system which parts of the ontology are definitely correct (and thus are a part of the background knowledge). This task can be accomplished using the functionality at the right-most side of the user interface of OntoDebug shown in Figure \ref{fig:koala:session}. In this example, the user works on problems of the ``Koala'' ontology of the \protege project, an ontology that was created for educational purposes which contains typical problems that can occur during ontology development. Specifically, in the example, the user has declared among other things that the axiom ``\emph{BA} (bachelor of arts)
%degree
is of type \emph{Degree}'' is definitely correct.

Once this optional step is done, the user can start the model-based debugging process. To this end, the tool, as mentioned above, supports two general strategies.

\begin{itemize}[leftmargin=12pt]
	\item First, the user can inspect the list of diagnoses returned by OntoDebug to locate the fault and add additional test cases if the list of diagnoses contains too many elements. Generally, the idea is that the provision of additional, carefully designed test cases will help to narrow down the set of
	possible diagnoses, i.e., the possible causes for the problems in the ontology.
	In the example shown in Figure \ref{fig:koala:session}, the user has specified one positive test case (``\emph{Student} is a subclass of \emph{Person}'') and a negative one (``\emph{Person} is a subclass of \emph{Marsupials}''), using the sub-window in the middle of the screen.
	
	\item The second supported debugging strategy is the query-based one. In this case, the tool will---based on the inconsistent (incoherent) ontology or the failing test cases---compute the first query to the user. In our example, the system determined a query consisting of two axioms shown in the top-left sub-window of the user interface. The two axioms to be categorized by the user are ``\emph{KoalaWithPhD} is a subclass of \emph{Koala}'' and ``\emph{KoalaWithPhD} is a subclass of \emph{Person}.'' The user can answer the query by using the green and red plus and minus symbols (or leave some axioms uncategorized), and then submit the answer to the system. The system adds the user's feedback to the ``Acquired Test Cases'' and then restarts the computations using the additionally provided information. In case the information was sufficient to identify a single diagnosis as the cause of the problem, the user is pointed to the faulty parts of the ontology. Otherwise, the system computes a new query to the user and the cycle repeats until only one diagnosis remains.
\end{itemize}

Generally, one main difference is that in the approach based on test cases the users have to think by themselves about good test cases, while in the case of interactive debugging, user responses to the system-generated queries are taken as additional test cases. In this latter case, the selection of the query, and correspondingly the test case(s), is based on an internal reasoning process that ensures that the most informative queries are chosen.

\section{Research Questions}
\label{sec:research-questions}
The main promise of interactive, query-based approaches is that they are able to systematically guide users (e.g., knowledge engineers or domain experts) through the debugging process and that after the interactive process the true cause of the observed discrepancies is found.
In contrast, there is limited support for users in the more traditional model-based debugging setting, where the users have to provide test cases manually in order to incrementally narrow down the set of fault candidates.

As discussed in Section \ref{sec:related-work}, computational analyses---such as measurements of time or an analysis of the number of required queries---can be insufficient to inform us about the usefulness  and acceptance of the corresponding tools, and
%, as such measurements are based on certain assumptions about the ability of an oracle to answer queries.
% PR: We can think of various other assumptions, NOT only the ability of users to answer queries...therefore i deleted this part -- apart from that, the sentence starts with ``As discussed before...'' which is why we do not need to discuss it again.
%Such measurements also
cannot tell us in which ways query-based debugging is advantageous over a test case based approach.

To address these open questions, we conducted a number of controlled (laboratory) studies, mainly consisting of ontology debugging exercises. We focus on the following main research questions in the context of model-based debugging:

\begin{enumerate}[label=RQ\arabic{*}, ref=Q\arabic{*}, leftmargin=3.0em]
	\item \label{item:RQ1} Is the debugging process more \emph{effective} when users are supported by a query-based debugging tool than when test cases are the only means to locate faults?
	\item \label{item:RQ2} Is the process more \emph{efficient} when users are supported by a query-based debugging tool?
	\item \label{item:RQ3} To what extent do the assumptions of MBD debugging techniques hold?
    \begin{enumerate}[label=RQ3.\arabic{*}, ref=Q\arabic{*}, leftmargin=3.0em]
        \item \label{item:RQ3.1} For the case of approaches based on test cases and candidate ranking% KS ranking is not introduced
        : Do users have ``perfect bug understanding'', i.e., do they reliably recognize the true cause of a discrepancy within a list of diagnoses?
        \item \label{item:RQ3.2} For the case of the query-based approach: Do users make errors when acting as oracles?
    \end{enumerate}
\end{enumerate}

\noindent The following studies were designed and executed.
\begin{itemize}
\item In our \emph{preliminary} study (\emph{Study 1}), our goal was to gauge the general usefulness of a test case based debugging approach. We specifically also explored the importance of the ranking of the fault candidates in this experiment (RQ3.1). The study also served us to further improve the design of the main study (\emph{Study 2}).
\item In \emph{Study 2}, we investigated the effectiveness and efficiency of the query-based and the test case based debugging approach (RQ1 and RQ2). In that context, we also examined the question of oracle errors (RQ3.2).
\end{itemize}

Additional pen-and-paper studies were conducted in the context of both \emph{Study 1} and \emph{Study 2} with the goal of deepening our understanding of the (types of) errors that occur while debugging. These insights are then used to devise a heuristic prediction model for such errors (RQ3.2). We discuss \emph{Study 1} in Section~\ref{sec:study1}, \emph{Study 2} in Section~\ref{sec:study2}, and the additional studies in Section~\ref{sec:prediction-model}.\footnote{The (anonymized) raw data obtained throughout \emph{Study 1} and \emph{Study 2} as well as the ontologies used in the experiments can be downloaded from \url{http://isbi.aau.at/ontodebug/evaluation}.}

\section{\emph{Study 1}: Investigating MBD-debugging With Test Cases}
\label{sec:study1}
\subsection{Design of the Pre-Study}
\subsubsection{Task}
The task of the participants in this study was to \emph{find the faulty axioms} (true diagnosis) in a given faulty ontology \begin{enumerate*}[label=\textit{(\roman*)}]
	\item based on a provided description of the \emph{intended} ontology in natural language
	\item using the OntoDebug tool described above
	\item by creating test cases manually (the query-based debugging functionality was not available to the users).
	
	The participants were explicitly instructed to
	\item constantly inspect the list of possible diagnoses throughout the debugging session and to
	\item mark the true diagnosis once they detected it in the list.
\end{enumerate*}
After a diagnosis was marked, the debugging session ended. In Figure~\ref{fig:koala:session}, the list of diagnoses is shown in the bottom-left sub-window
labeled with ``Possible Ontology Repairs''.

\subsubsection{Ontologies}
In order to make sure that the outcomes regarding the usefulness of the test case based debugging approach do not depend on the specifics of a certain ontology, two different ontologies describing two different domains were used in the study. The first one corresponded to a (simplified) model of our university %the ``University of Klagenfurt''
(\emph{university} domain) and the second one was a real-world knowledge base made available to us by the ``Communal IT Center of Carinthia'' (\emph{IT} domain). We prepared the ontologies for the study by injecting five faults into each of them such that the resulting ontologies were inconsistent and incoherent in themselves. That is, for both ontologies the true diagnosis included five faulty axioms (as shown in Table \ref{tab:faults}).
The designed ontologies were similar in size and complexity. For example, both included about 50 classes, 90 subclass relationships, and 20 object properties. Moreover, both included roughly equally complex logical formalisms and used the full expressivity of the Description Logic $\mathcal{SROIQ}$ \cite{DLHandbook,horrocks2006even} or, respectively, OWL 1.1 \cite{owl1.1_spec}.

\begin{table}[h!t]
	\footnotesize
	\caption{Faulty ontology axioms (university domain) in OWL Manchester Syntax \cite{horridge2006manchester}.}
%	Language keywords are printed in bold face.}
	\label{tab:faults}
	\centering
	\begin{tabular}{|l|p{7.5cm}|}
		\hline
		Nr. & Faulty Axiom\\ \hline
		1 & \stt{\scriptsize Department SubClassOf offers only Course} \\
		2 & \stt{\scriptsize Library SubClassOf offers only Visitation} \\
		3 & \stt{\scriptsize Research\_Event SubClassOf has\_Speaker only (Person and \mbox{~~~~~}(has\_Degree some Degree))} \\
		4 & \stt{\scriptsize Assembly\_Hall DisjointWith Room} \\
		5 & \stt{\scriptsize Department DisjointWith Room} \\
		\hline
	\end{tabular}
\end{table}

\subsubsection{Participants}
We recruited 29 participants for the study. All participants were computer science students of our university and were enrolled in an ongoing master program course on knowledge engineering. During this course, the participants, who already had a background in logics, were introduced to model-based debugging, formal ontologies, Description Logics, and the OWL language. The participants also had first experiences in designing ontologies with \protege and debugging them with OntoDebug. Overall, the participants were very homogeneous with respect to their knowledge and background.

\subsubsection{Independent Variables}
We considered two independent variables,
	the ontology to be debugged (\emph{university} vs.\ \emph{IT}) and
	the position (\emph{visible} vs.\ \emph{not visible}) of the true diagnosis in the list of diagnoses returned by the debugger.
Each participant was randomly assigned to one ontology and one of two configurations regarding the position of the true diagnosis.

Similar to the work in \cite{Parnin:2011:ADT:2001420.2001445}, we varied the position to assess the importance of the ranking of the diagnoses returned by the system. Specifically, in the \emph{visible} case, the true diagnosis, which comprised all actually faulty axioms of the ontology, was placed within the top three diagnoses and was therefore always visible to the user. In the other case (\emph{not visible}), the true diagnosis was further down the list.
Generally, the diagnosis problem was designed in a way that the initial list of diagnoses (before further test cases are specified) is comparably large, including over 150 diagnoses in each case.
%
%\todost{Go through the entire paper and unify axioms/statements/definitions/sentences/etc.\ as discussed -- please help!}
%\todost{What I find misleading: sometimes we talk about \emph{true diagnosis} vs.\ \emph{diagnoses} (i.e.\ tacitly meant are the ``wrong'' diagnoses) and sometimes we use \emph{diagnosis candidates} to refer to the ``wrong'' diagnoses -- please help fix that!}
%\todost{target diagnosis vs\ actual diagnosis vs.\ true diagnosis -- check the entire paper and unify.}
%\tododj{I agree, please fix. I added informal definitions of the concepts of target diagnosis and candidates in the section on related works.}

\subsubsection{Dependent Variables}
\label{sssec:dependent_vars_study1}
We made a variety of automated, \emph{objective} measurements while the participants were executing the task, like the needed \emph{time}, the \emph{number of user interactions} (mouse clicks) in the debugger, and the \emph{number of diagnoses still in the list} of diagnoses when the participants
submitted the diagnosis which they thought is the correct one.
In the context of \emph{Study 1} the most important automated measurement was on the \emph{correctness} of the debugging process in terms of
\begin{enumerate*}[label=(\textit{\roman*})]
	\item the fraction of correctly identified faulty axioms and
	\item the fraction of users who correctly identified all five faulty axioms (i.e.\ the true diagnosis).
\end{enumerate*}

Moreover, the participants had to specify their \emph{subjective} degree of belief (\emph{confidence}) in having solved the fault localization task correctly. For this, they were instructed to use a range between 0 (certain that the marked diagnosis \emph{is not} the true one) and 100 (certain that the marked diagnosis \emph{is} the true one).

\subsection{Experiment Execution}
\label{sssec:experiment_execution_study1}
The study was conducted in one of the computer labs of our university. The required software was pre-installed on the lab computers. All of the computers were identically equipped.
After being informed about the tasks of the study and after the participants had declared their consent, they were provided with detailed material on paper. The handout essentially included a description of the domain that was incorrectly modeled by the ontology the participants had to debug. Thus, the paper characterized the \emph{intended ontology} as discussed in Section \ref{sec:kb-debugging}.

The description was given as a natural language text, with important concepts highlighted. In particular, class and property names in the ontology were \emph{italicized} and \underline{underlined}, respectively. An example of such a description from the university domain is the following:

\begin{myquote}\small
	From an organizational point of view, the University is subdivided into several \emph{OrganizationalUnits}. Each \emph{OrganizationalUnit} \underline{employs} some \emph{OfficeEmployee}(s) and some \emph{Teacher}(s), \underline{has} some \underline{\emph{Room}}(s) which is/are (an) \emph{Office}(s), \underline{is directed by} exactly one \emph{Director} and \underline{is located in} some \emph{Building}. Two special types of \emph{OrganizationalUnits} are the \emph{Directorate} and the \emph{HumanResourcesUnit}.
\end{myquote}

Before the participants started their task, they received
another brief tutorial on
how to debug an ontology with the OntoDebug tool. They used %a different ontology.
the ``Koala'' ontology that is available in \protege (cf.\ Sec.~\ref{sec:ontodebug_plugin}) for that purpose.
During the experiment, the participants were not allowed to talk to each other. The participants were supervised by three instructors, who were present to answer questions in case of problems with the software.

\subsection{Outcomes of Study 1}
The measurements obtained in \emph{Study~1} are summarized by Figures~\ref{fig:study1_overview} and \ref{fig:study1_vioplots}.
As mentioned above, the main question of this pre-study was 
\begin{enumerate*}[label=\textit{(\roman*)}]
	\item to gauge the general usefulness of MBD-debugging with test cases and 
	\item to assess the importance of the ranking of the diagnoses. Furthermore, a side goal was to obtain experiences regarding the study design for the main study (\emph{Study 2}).
\end{enumerate*}
%We discuss the results in what follows.
% TODO: add the contingency table here. there is some issue with the formatting ..

%\subsubsection{General Usefulness of MBD-based Debugging}
\subsubsection{General Usefulness of Model-based Debugging}
\label{sssec:usefulness_of_MBD_debugging}
\begin{figure}
	\centering
	\footnotesize
	\includegraphics[width=0.99\linewidth]{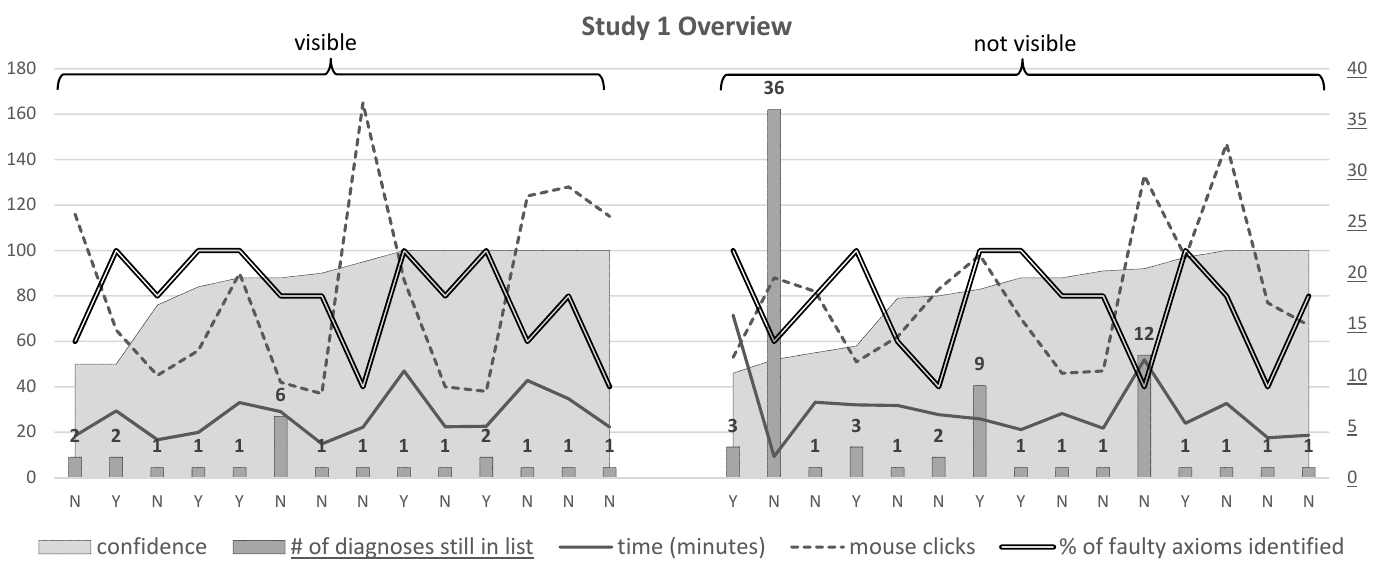}
	%figs/study1--overview1_CUT
	\caption{Overview of the outcomes of \emph{Study~1}. The figure shows the measurements for the dependent variables for all 29 debugging sessions, grouped by the position (``visible'' left, ``not visible'' right) of the true diagnosis in the diagnoses list, and sorted from low to high confidence. The labels along the x-axis indicate whether the true diagnosis was found (``Y'') or not (``N'') during the respective session. Variables plotted w.r.t.\ the right y-axis are underlined. The numbers (ranging from 1 to 36) in the plot indicate the exact value of the ``\# of diagnoses still in list'' variable.}
	\label{fig:study1_overview}
\end{figure}

\begin{figure}[ht]
	\label{ fig7}
	\begin{minipage}[b]{0.25\linewidth}
		\centering
		\includegraphics[width=.75\linewidth]{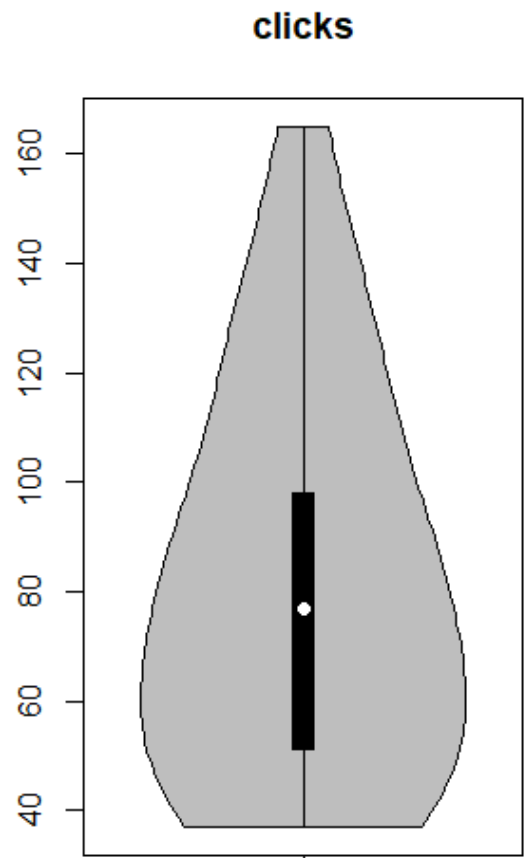}
%		\caption{Initial condition}
%		\vspace{4ex}
	\end{minipage}%%
	\begin{minipage}[b]{0.25\linewidth}
		\centering
		\includegraphics[width=.745\linewidth]{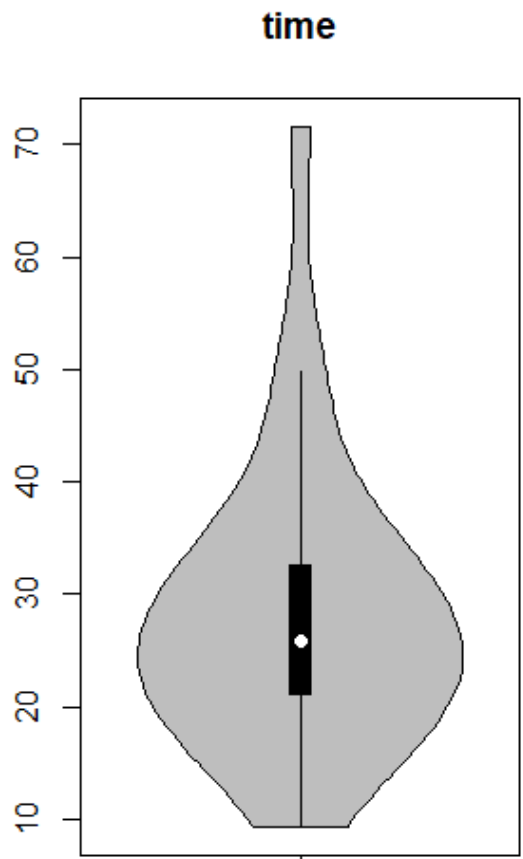}
%		\caption{Rupture}
%		\vspace{4ex}
	\end{minipage}%%
	\begin{minipage}[b]{0.25\linewidth}
		\centering
		\includegraphics[width=.76\linewidth]{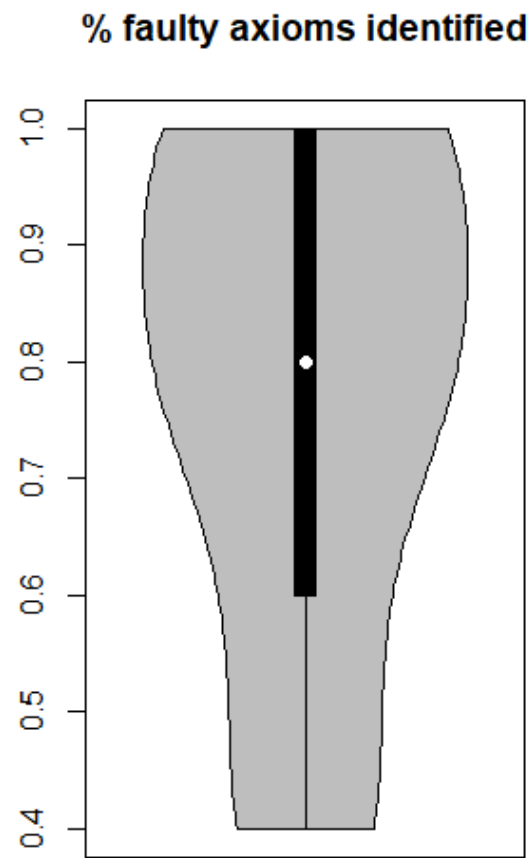}
%		\caption{DFT, Initial condition}
%		\vspace{4ex}
	\end{minipage}%%
	\begin{minipage}[b]{0.25\linewidth}
		\centering
		\includegraphics[width=.78\linewidth]{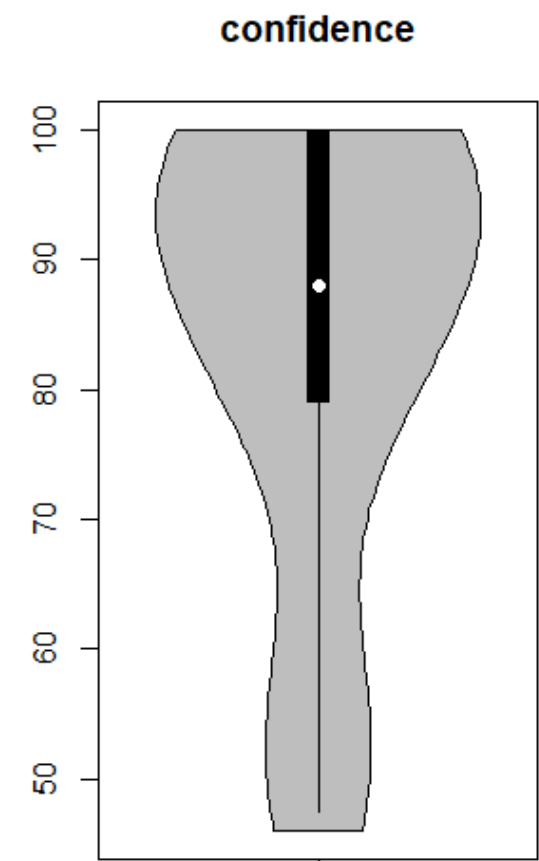}
%		\caption{DFT, rupture}
%		\vspace{4ex}
	\end{minipage}
\caption{Violin plots showing the distribution of the dependent variables in \emph{Study~1}.}
\label{fig:study1_vioplots}
\end{figure}

On average, the participants took about 28 minutes and 81 mouse clicks for the task before they submitted their solution.\footnote{The standard deviation was 12.6 minutes (time) and 35 mouse clicks, respectively.}
Overall, the participants correctly identified as many as 77\,\% of the problematic axioms,\footnote{The standard deviation was 21\,\%.}
%TODO Should we really mention the std dev regarding the % here?
i.e., almost four out of the five injected faults shown in Table \ref{tab:faults} were eliminated. From the 29 participants, 10 (34.4\,\%) correctly identified the true diagnosis, i.e., all five faulty axioms (cf.\ Figure~\ref{fig:study1_overview}).\footnote{Note, in Figures~\ref{fig:study1_overview} and \ref{fig:study2_overview} we use line and area charts (instead of, e.g., bar charts or dot plots) for reasons of clarity and better distinguishability between the plotted variables, although the x-axis is in fact a categorical axis.}

Overall, we find this result very positive, given the complexity of the task. The study clearly indicates that
%MBD-based
model-based
debugging is actually helpful for knowledge engineers. Since we did not observe any statistically significant differences between the observations that were made for two different ontology debugging problems (\emph{university} and \emph{IT}), we are confident that the usefulness of the approach is not limited to just one domain.

There were various reasons why some participants did not successfully find all faults. A main issue appeared to be a certain lack of attentiveness and precision when reading the natural language specification of the intended ontology.
Based on these observations, we revised some of the specifications, e.g., by removing possible ambiguities, when designing \emph{Study 2}.
To a certain extent, it also seemed that some participants did not properly understand the semantics of certain elements of the knowledge representation language.

\subsubsection{Importance of Ranking of Candidates (RQ3.1)}
In the context of RQ3.1,
our goal was to investigate if the capability of a debugger to rank the true diagnosis higher in a list of candidates directly translates into a more effective debugging process. Table~\ref{tab:position_of_true_diag_vs_correctness} shows in how many cases the true diagnosis---which comprises all five injected faults---was found, depending on whether it was among the top-ranked (visible) candidates or not.

\begin{table}[h!t]
	\footnotesize
	\centering
	\caption{Relationship between full correctness of the debugging task and visibility of the true diagnosis in the list of diagnoses presented to the participant.}
	\label{tab:position_of_true_diag_vs_correctness}
	\begin{tabular}{ll|l|l|}
		\cline{3-4}
		&     & \multicolumn{2}{l|}{\textbf{true diagnosis visible}} \\ \cline{3-4}
		&     & yes                   & no                  \\ \hline
		\multicolumn{1}{|l|}{\multirow{2}{*}{\begin{tabular}[c]{@{}l@{}}\textbf{true diagnosis found}\end{tabular}}} & yes & 5                     & 5                   \\ \cline{2-4}
		\multicolumn{1}{|l|}{}                                                                                   & no  & 9                     & 10                  \\ \hline
	\end{tabular}
\end{table}

Interestingly, the observations shown in Table~\ref{tab:position_of_true_diag_vs_correctness} do not provide evidence that the users were more effective when the true diagnosis was always visible.\footnote{
	%	What is already obvious from Table~\ref{tab:position_of_true_diag_vs_correctness},
	This is supported by Fisher's Exact Test \cite{bhattacharyya1977} (p-value = 1.00).}
%\footnote{Albeit obvious from Table~\ref{tab:position_of_true_diag_vs_correctness}, a verification using Fisher's Exact Test \cite{bhattacharyya1977} yields a p-value = 1.00.}
%\footnote{As verified by a Chi-Squared Test \cite{bhattacharyya1977} (p-value = 0.89).}
Such a non-effect of varying the position of the fault in a ranked list was also reported in \cite{Parnin:2011:ADT:2001420.2001445}.

Moreover, 10 of the 14 participants of the group where the true diagnosis was ranked highly
continued specifying test cases until only one diagnosis was left in the list (cf.\ Figure~\ref{fig:study1_overview})---even though all participants were explicitly instructed to constantly inspect the list of diagnoses and mark the true diagnosis once they detected it in the list.
A large number of participants therefore did not recognize the actual fault even though it was shown to them.

These findings challenge the assumption of a ``perfect bug understanding'' of the users, i.e., they do \emph{not} always immediately identify a fault when they are pointed to it. In other words, even if the true diagnosis was visible to the participants, they \begin{enumerate*}[label=\textit{(\roman*)}] \item did not recognize it in the majority of the cases and \item did not identify it more often than other participants to which the true diagnosis was not (always) visible.\end{enumerate*} As a result, fault ranking metrics should not be considered as the only measure when different algorithmic debugging strategies are compared \cite{Parnin:2011:ADT:2001420.2001445}.

\subsubsection{Additional Observations (Study 1)}
\label{sssec:additional_observations_study1}

\noindent\emph{Positive test cases are more reliable:} From the 244 test cases provided by the participants (8 on average per debugging session), the majority (71\,\%) were \emph{positively} formulated, i.e., they described required entailments. The participants therefore seemed to feel more comfortable specifying things that must be entailed than those that must not. An analysis of the fault rates for positive and negative test cases indeed confirmed that negative ones, i.e., formulated non-entailments, were significantly\footnote{According to a
%	Chi-Squared Test with $\alpha = 0.01$ (p-value = 0.00496) as well as a
(two-tailed) Fisher's Exact Test with $\alpha = 0.01$ (p-value = 0.008).}
%TODO should we discuss the fixed row and columns sums that are assumed by Fisher Test? Normally, literature suggests that it is OK that only EITHER row sums OR column sums are fixed -- as in our case -- and if not both are fixed, the test just becomes more conservative (i.e. loses power), however we can highly significant results suggesting the rejection of null hypo EVEN THOUGH it might be too conservative, i.e. in favor of the null hypo
more often faulty (24\,\% vs.~10\,\%, see Table~\ref{tab:type_of_testcase_vs_faultiness}). This result suggests that it can be better to ask users questions with a bias towards
%\tododj{What is a bias toward the positive answer? Who/what is biased?}
the positive answer\footnote{A ``bias towards the positive answer'' means that the estimated probability of getting a positive answer to the question is higher than that of a negative answer.} in query-based KB debugging, in order to minimize oracle errors.

\begin{table}[h!t]
	\footnotesize
	\centering
	\caption{Relationship between the type of formulated test case and its faultiness.}
	\label{tab:type_of_testcase_vs_faultiness}
	\begin{tabular}{ll|l|l|}
		\cline{3-4}
		&     & \multicolumn{2}{l|}{\textbf{type of test case}} \\ \cline{3-4}
		&     & positive                   & negative                  \\ \hline
		\multicolumn{1}{|l|}{\multirow{2}{*}{\begin{tabular}[c]{@{}l@{}}\textbf{test case faulty}\end{tabular}}} & yes & 18                     & 17                   \\ \cline{2-4}
		\multicolumn{1}{|l|}{}                                                                                   & no  & 156                     & 53                  \\ \hline
	\end{tabular}
\end{table}

\noindent\emph{Users can be overconfident:} The participants of the study were, on average, overconfident (cf.\ Figure~\ref{fig:study1_overview}). That is, the average confidence value expressed by the participants regarding the correctness of the identified diagnosis was at about 83\,\% (cf.\ Figure~\ref{fig:study1_vioplots})\footnote{In all the presented violin plots, the white dot indicates the median and the thick box ranges from the first to the third quartile.} whereas only 34\,\% of them have correctly located the true diagnosis. In other words, the average self-reported confidence of users in their own success overestimates the actual user success rate.
%\footnote{We are aware that the two measures of this comparison---average user confidence in the correctness of the result and proportion of users who did achieve the correct result---cannot be directly compared. However, it is interesting to observe that the two measurements show a certain discrepancy.}
%\footnote{We are aware that the two measures of this comparison---fault identification rates and self-reported confidence values---cannot be directly compared. However, it is interesting to observe that the two measurements show a certain discrepancy.}
%(see Sec.~\ref{sssec:usefulness_of_MBD_debugging}).
Interestingly, the confidence of those participants who did not find the true diagnosis was even slightly higher than the confidence of the successful participants.

Overall, this can be seen as an indicator that subjective confidence estimates have to be handled with care \cite{Rodler2013} when they are intended to be used to guide the debugging process \cite{Shchekotykhin2012}.

\noindent\emph{Users consider themselves as imperfect oracles:}
%We found that 69\% of all users, and even 80\% of the successful ones, were not fully confident about the correctness of their actions.
We found that only 31\,\% of all users, and an even lower 20\,\% of the ones that successfully found all faults, were \emph{fully} confident
about the correctness of their debugging actions (cf.\ Figure~\ref{fig:study1_overview}).
This teaches us that humans generally
do not regard themselves as perfect oracles for knowledge engineering tasks, which questions the frequently made ``perfect oracle'' assumption. We pick up on this discussion again in Sec.~\ref{sssec:existence_of_oracle_errors} and Sec.~\ref{sec:prediction-model}.

\noindent\emph{Completion time and user activity as success predictors (cf.\ Figure~\ref{fig:study1_overview}):} Participants who correctly identified the true diagnosis required on average more time (33 minutes) and specified more test cases (10). However, they needed fewer interactions (71 clicks) than those who submitted a wrong diagnosis (26 minutes, 8 test cases, 87 clicks). This indicates that successful users worked more thoroughly and were more persistent in their testing activity. Unsuccessful ones, in contrast, required more interactions as they more frequently edited, deleted or re-added test cases.
An atypically high editing activity can thus be considered as an indicator that a user requires more assistance for the given task.

\section{\emph{Study 2}: On the Usefulness of Query-based Debugging}
\label{sec:study2}
Having established that
%MBD-based
model-based
debugging leads to a good debugging performance, the goal of \emph{Study 2} was to answer our main research questions RQ1 and RQ2 on the efficiency and effectiveness of query-based debugging as opposed to a test case based approach. In other words, do users need less time/effort when supported by a query-based debugger (efficiency) and do they find more faults (effectiveness)?

\subsection{Design of the Study}
\subsubsection{Task}
As in the pre-study, the general task of the participants was to find the actually faulty axioms (true diagnosis) in given faulty ontologies
\begin{enumerate*}[label=\textit{(\roman*)}]
	\item based on a description of the \emph{intended} ontology in natural language
	\item using the OntoDebug tool. However, now
	\item every participant had to debug two ontologies,
%	\item
	one using the query-based and
%	\item
	the other using the test case based approach.
\end{enumerate*}

\subsubsection{Ontologies}
Similar ontologies were used as in the pre-study---one describing a \emph{university}, and one describing an \emph{IT} domain, and both again corresponding to the Description Logic $\mathcal{SROIQ}$.
%Also like in \emph{Study 1}, we
We prepared the ontologies for the study by injecting a number of faults into each of them, leading to both inconsistency and incoherency, similarly as in \emph{Study~1}. However, the ontologies were roughly 20\,\% larger in terms of their size (e.g., number of axioms and classes) than the ones used in \emph{Study 1}; still, the size and complexity of both ontologies was
roughly equal.
The ontologies were enlarged to achieve a higher number of fault candidates. Concretely, the size of the initial list of diagnoses for both ontologies was now over 1000.
This made the diagnosis problems objectively harder than in the pre-study. The reason for this was to compensate for the lower number of participants (23)
in \emph{Study 2}, which makes it more challenging to achieve statistically significant results. In fact, if any effects (e.g., regarding time or user interactions) of employing the query-based debugging method can be found, then they are likely to be larger for harder debugging problems.

\subsubsection{Participants}
For \emph{Study 2}, we could draw on 23 participants. Again, all of them were attendees of a university master program course on knowledge engineering. However, the focus of the course was now shifted towards Semantic Web technologies to achieve a better preparation of the students for the study.
As a consequence, the participants of \emph{Study 2} had a better education on model-based diagnosis, formal ontologies, ontological reasoning, and the used knowledge representation language than those of \emph{Study 1}. Moreover, they had more experience with \protege and OntoDebug.

\subsubsection{Independent Variables}
The two independent variables we used were
the ontology to be debugged (\emph{university} vs.\ \emph{IT}) and
the debugging strategy (\emph{query-based} vs.\ \emph{test case based}).
We used a within-subjects experiment design in this study, which involves each participant consecutively working on both ontologies and consecutively using both debugging strategies. Thus, we randomly assigned each participant to one of the following configurations:\footnote{The random variation of the order of the tasks is important to avoid systematic learning effects.}
\begin{itemize}[noitemsep,leftmargin=16pt]
	\item Task 1: \emph{university} with \emph{queries}. Task 2: \emph{IT} with \emph{test cases}.
	\item Task 1: \emph{university} with \emph{test cases}. Task 2: \emph{IT} with \emph{queries}.
	\item Task 1: \emph{IT} with \emph{queries}. Task 2: \emph{university} with \emph{test cases}.
	\item Task 1: \emph{IT} with \emph{test cases}. Task 2: \emph{university} with \emph{queries}.
\end{itemize}

\subsubsection{Dependent Variables}
In terms of measurements, we recorded the same aspects as in \emph{Study 1} (see Section~\ref{sssec:dependent_vars_study1}), i.e., \emph{time}, \emph{number of user interactions}, \emph{number of diagnoses still in the list}, \emph{correctness} (fraction of faulty axioms found, fraction of users finding the true diagnosis), and \emph{confidence}.

\subsection{Experiment Execution}
The experiment execution was exactly the same as in \emph{Study~1}, see Section~\ref{sssec:experiment_execution_study1}.

%\begin{figure}[ht]
%	\begin{minipage}[b]{0.5\linewidth}
%		\centering
%		\includegraphics[width=.75\linewidth]{figs/s2_test-cases_fraction_correct_ax_CUT}
%		%		\caption{Initial condition}
%		%		\vspace{4ex}
%	\end{minipage}%%
%	\begin{minipage}[b]{0.5\linewidth}
%		\centering
%		\includegraphics[width=.745\linewidth]{figs/s2_queries_fraction_correct_ax_CUT}
%		%		\caption{Rupture}
%		%		\vspace{4ex}
%	\end{minipage}%%
%	\caption{Vioplots showing the distribution of the percentage of correctly identified faulty axioms in \emph{Study~2}.}
%	\label{fig:study2_vioplots}
%\end{figure}

\subsection{Outcomes of Study 2}

\subsubsection{Effectiveness of Query-based Debugging (RQ1)}
\begin{figure}[t]
	\centering
	\footnotesize
	\includegraphics[width=0.99\linewidth]{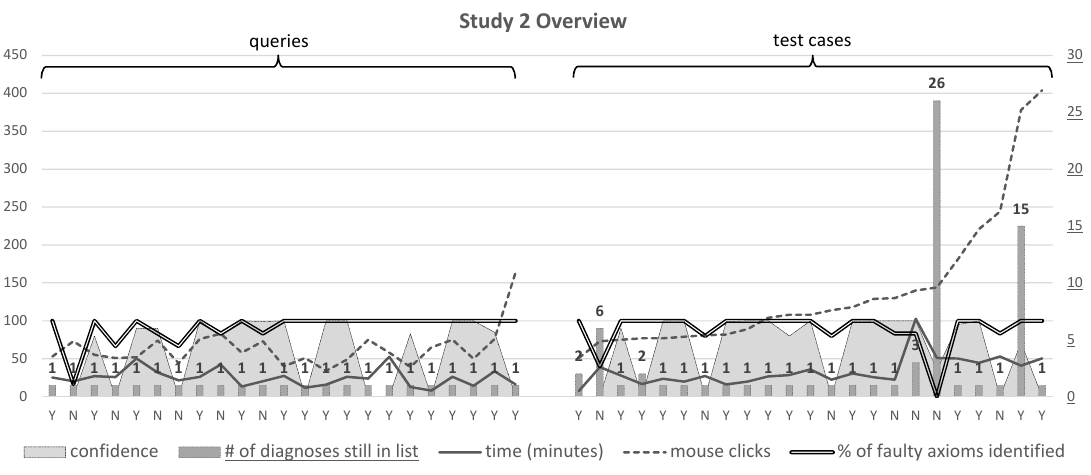}
	% figs/study2--overview23_CUT
	\caption{Overview of the outcomes of \emph{Study~2}. The figure shows the measurements for the dependent variables for all 46 debugging sessions (2 per user), grouped by the used debugging strategy (``queries'' left, ``test cases'' right;
		%	$i$-th x-axis entry on the left refers to the same user as $i$-th x-axis entry on the right
		$i$-th x-axis entry starting from the left in the ``queries'' block refers to the same user as $i$-th x-axis entry starting from the left in the ``test cases'' block). Records are sorted by the number of mouse clicks of the ``test cases'' sessions from low to high. The labels along the x-axis indicate whether the true diagnosis was found (``Y'') or not (``N'') during the respective session. Variables plotted w.r.t.\ the right y-axis are underlined. The numbers (ranging from 1 to 26) in the plot indicate the exact value of the ``\# of diagnoses still in list'' variable.
		%	Note, the data for the debugging sessions of one and the same user appear at one and the same
	}
	\label{fig:study2_overview}
\end{figure}

\begin{figure}[t]
	\centering
	\footnotesize
	\includegraphics[width=.8\linewidth]{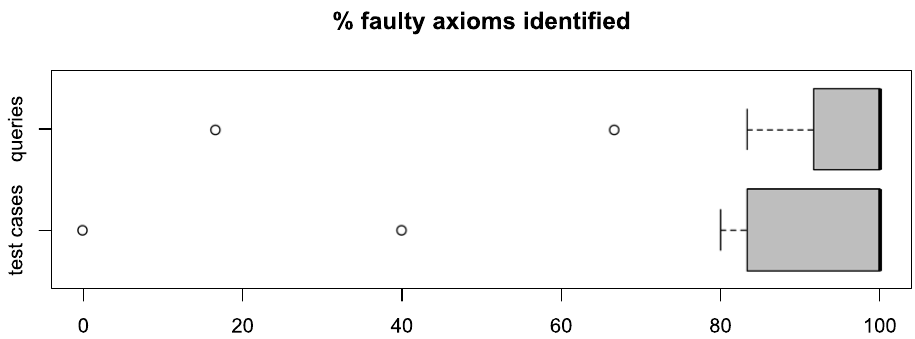}
	\caption{Boxplots showing the distribution of the \% of identified faulty axioms per debugging session in \emph{Study~2} for the query-based vs.\ the test case based approach.}
	\label{fig:study2_boxplots_fraction_faulty_ax_found}
\end{figure}

To assess the effectiveness of the two debugging strategies, we analyzed how many of the faulty axioms were successfully identified by the participants. Across both ontologies, the participants on average found 91.3\,\% of the faults when they were supported by the query-based debugger and 89.1\,\% when the debugging process was based on test cases (as in \emph{Study 1}).\footnote{The standard deviation amounts to 19\,\% (queries) and 23\,\% (test cases).} Figures~\ref{fig:study2_overview} and \ref{fig:study2_boxplots_fraction_faulty_ax_found} show the (distribution of the) achieved success rates for both debugging techniques.
The differences were not statistically significant. We therefore conclude that in this experiment the query-based approach did \emph{not} further increase the effectiveness of the debugging process.

Note, however, that in both cases the success rate was higher than in \emph{Study~1}, where 77\,\% of the faults were identified by the participants. We attribute this to the fact that---based on the learnings from \emph{Study 1}---we were more successful in motivating the participants to work more carefully. In addition, the participants of \emph{Study 2}
were, as mentioned, better trained in ontology
engineering than those of \emph{Study 1}.  As a result, it became difficult to greatly increase the already high success rate (89.1\,\%) obtained by participants who relied on test case based debugging.

Like in \emph{Study 1}, we also analyzed how many of the participants could correctly identify \emph{all} faulty axioms (i.e.\ the true diagnosis) in each ontology. We again found no statistically significant difference between the two debugging approaches (cf.\ Table~\ref{tab:debugging_method_vs_correctness}). Generally, across the ontologies, the fraction of fully successful trials was much higher than in \emph{Study 1}. About 72\,\% of the participants were able to find all problems in the respective ontologies. Interestingly, we registered a non-conformity between the
two ontologies this time. Over 85\,\% of the participants were able to find all faults in the \emph{university} ontology, with no significant differences with respect to the debugging method.
However, in the \emph{IT} domain, only 57\,\% were fully successful. A potential reason for this result could lie in the prior knowledge of the participants with regard to the two domains. More research is however required to better understand this phenomenon.

\begin{table}[h!t]
	\footnotesize
	\centering
	\caption{Relationship between the used debugging approach and the success in finding the true diagnosis.}
	\label{tab:debugging_method_vs_correctness}
	\begin{tabular}{ll|l|l|}
		\cline{3-4}
		&     & \multicolumn{2}{l|}{\textbf{debugging approach}} \\ \cline{3-4}
		&     & queries                   & test cases                  \\ \hline
		\multicolumn{1}{|c|}{\multirow{2}{*}{\parbox{3cm}{\textbf{true diagnosis found}}}} & yes & 17                     & 16                   \\ \cline{2-4}
		\multicolumn{1}{|l|}{}                                                                                   & no  & 6                     & 7                  \\ \hline
	\end{tabular}
\end{table}

%  \begin{minipage}{\textwidth}
%	\begin{minipage}[b]{0.37\textwidth}
%		\centering
%		\includegraphics[width=.78\linewidth]{figs/s2_frac_faulty_ax_1_CUT}
%		\captionof{figure}{bla bla}
%	\end{minipage}
%	\hfill
%	\begin{minipage}[b]{0.63\textwidth}
%		\footnotesize
%		\centering
%		\begin{tabular}{ll|l|l|}
%			\cline{3-4}
%			&     & \multicolumn{2}{l|}{\textbf{debugging approach}} \\ \cline{3-4}
%			&     & queries                   & test cases                  \\ \hline
%			\multicolumn{1}{|c|}{\multirow{2}{*}{\parbox{1.7cm}{\textbf{true diag-\\nosis found}}}} & yes & 17                     & 16                   \\ \cline{2-4}
%			\multicolumn{1}{|l|}{}                                                                                   & no  & 5                     & 6                  \\ \hline
%		\end{tabular}
%		\captionof{table}{Relationship between the used debugging approach and the success in finding the true diagnosis.}
%		\label{tab:debugging_method_vs_correctness}
%	\end{minipage}
%\end{minipage}

%\begin{figure}[ht]
%	\begin{minipage}[b]{0.5\linewidth}
%		\centering
%		\includegraphics[width=.75\linewidth]{figs/s2_test-cases_fraction_correct_ax_CUT}
%		%		\caption{Initial condition}
%		%		\vspace{4ex}
%	\end{minipage}%%
%	\begin{minipage}[b]{0.5\linewidth}
%		\centering
%		\includegraphics[width=.745\linewidth]{figs/s2_queries_fraction_correct_ax_CUT}
%		%		\caption{Rupture}
%		%		\vspace{4ex}
%	\end{minipage}%%
%	\caption{Vioplots showing the distribution of the percentage of correctly identified faulty axioms in \emph{Study~2}.}
%	\label{fig:study2_vioplots}
%\end{figure}

\subsubsection{Efficiency of Query-based Debugging  (RQ2)}
\begin{figure}[ht]
	\centering
	\footnotesize
	\includegraphics[width=.8\linewidth]{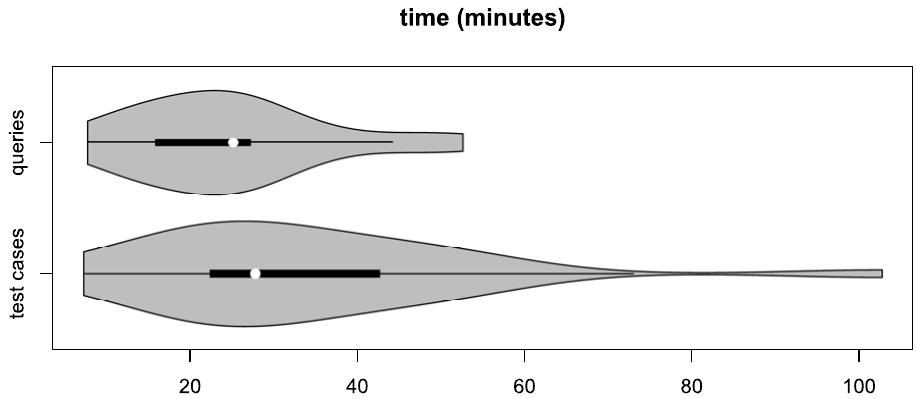}
	\caption{Violin plots showing the distribution of the debugging task completion times in \emph{Study~2} for the query-based vs.\ the test case based approach.}
	\label{fig:study2_vioplots_time}
\end{figure}

\begin{figure}[ht]
	\centering
	\footnotesize
	\includegraphics[width=.8\linewidth]{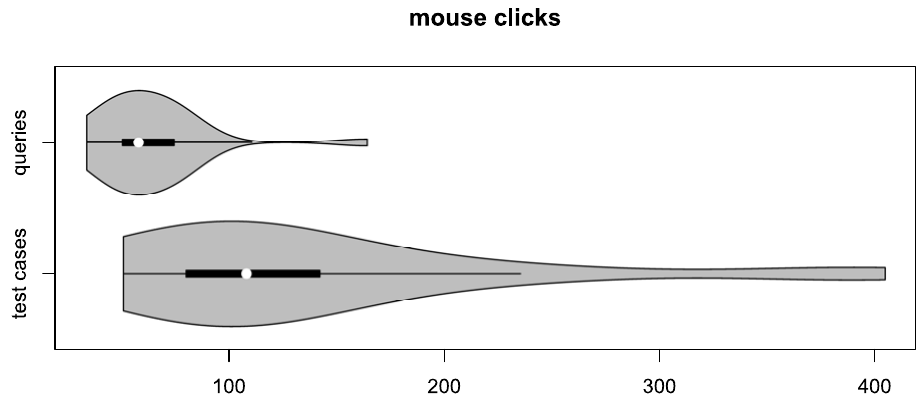}
	\caption{Violin plots showing the distribution of the number of user interactions to complete the debugging task in \emph{Study~2} for the query-based vs.\ the test case based approach.}
	\label{fig:study2_vioplots_clicks}
\end{figure}

To assess if the query-based debugging technique helps users to accomplish the debugging task faster and with less effort, we compared both the overall time needed by the participants and the number of required user interactions (mouse clicks) in the debugging tool across the two debugging strategies. The (distribution of the) time and user interaction measurements throughout \emph{Study~2} is summarized by Figures \ref{fig:study2_overview}, \ref{fig:study2_vioplots_time} and \ref{fig:study2_vioplots_clicks}.

Participants who were supported by the query-based debugging tool on average needed 24.9 minutes. When
using test cases without query support,
the average time was 34.0 minutes\footnote{The standard deviation comes to 11 minutes (queries) and 19 minutes (test cases).}, which amounts to an overhead of 37\,\%.
Looking at the number of required user interactions, the differences are even stronger. With the query-based debugging tool, the number of mouse clicks was more than halved and reduced from about 139 to 64 clicks on average.\footnote{Standard deviation: 25 clicks (queries) and 90 clicks (test cases).} The differences regarding both time and interactions were statistically significant according to a Wilcoxon Rank-Sum Test;\footnote{Since literature is not always consistent when referring to Wilcoxon's test(s), note that we stick to the description of the test(s) given in \cite{bhattacharyya1977}. Further note that Wilcoxon's Rank-Sum test compares \emph{independent} samples whereas Wilcoxon's Signed Rank test compares \emph{paired} data.}
in the case of time to the level $\alpha=0.05$ (p-value = 0.0418), and for clicks to the level $\alpha = 0.00001$ (p-value < 0.00001).\footnote{Also, when viewing the data as paired (each participant did use both queries and test cases, but each for a different ontology), the results in both cases are highly significant (for $\alpha = 0.05$ and $\alpha = 0.0001$, respectively).}

%When restricting the data to those cases representing fully
%%TODO adapt -- say ``almost''
%successful debugging sessions (true diagnosis found), the difference in the task completion time between query-based and test case based is even higher---and statistically significant as per a Wilcoxon Rank-Sum Test with $\alpha = 0.05$ (p-value = 0.0465).
%%\tododj{Das ist schon seltsam, dass hier zwei p-Werte vorkommen, die sich erst in der dritten Nachkommastelle unterscheiden und sehr knapp an 0.05 sind.}
%Particularly, we measured a time overhead of more than 30\,\% (33 vs.\ 25 minutes) to find \emph{all} faults when using test cases instead of queries.
%When restricting the data to those cases representing ``reasonable'' debugging attempts (i.e., at most one of the faulty axioms was not found), the difference in the task completion time between query-based and test case based is even higher---and statistically significant as per a Wilcoxon Rank-Sum Test with $\alpha = 0.05$ (p-value = 0.0465).
%%\tododj{Das ist schon seltsam, dass hier zwei p-Werte vorkommen, die sich erst in der dritten Nachkommastelle unterscheiden und sehr knapp an 0.05 sind.}
%Particularly, we measured a time overhead of more than 30\,\% (33 vs.\ 25 minutes) to find \emph{all} faults when using test cases instead of queries.

Overall, we conclude from the experiments that query-based debugging support is beneficial in terms of the efficiency of the debugging process.

\subsubsection{Additional Observations (Study 2)}
\label{sec:additional_obs_study2}
%\noindent\emph{Users feel equally confident using both debugging approaches:} While again overconfident in general (cf.\ Section~\ref{sssec:additional_observations_study1}), the participants were on average approximately equally confident (93\,\% in case of query assistance, 92\,\% when using test cases) about having made no mistakes in the debugging process at all.
%%, both when using queries and test cases.
\noindent\emph{Users feel equally confident using both debugging approaches:} While again overconfident in general (cf.\ Section~\ref{sssec:additional_observations_study1}), the participants were approximately equally confident about having made no mistakes in the debugging process at all, both when using queries and test cases.\footnote{Note that in Figure~\ref{fig:study2_overview} some confidence values seem to be zero. However, in fact, these cases represent \emph{unknown} confidence values where participants did not provide an answer to the question about their subjective belief in the correctness of their debugging result.} Specifically, the average confidence in case of query assistance was 93\,\% and 92\,\% when using test cases.\footnote{Standard deviation: 8\,\% (queries) and 17\,\% (test cases).}

\noindent\emph{Intuitive focus on mere query answering:}
Interestingly, without giving the participants who used the query-based debugger any instructions to do so, all of them continued answering queries until a single diagnosis was left (cf.\ Figure~\ref{fig:study2_overview}). Apparently, they therefore did not rely on the list of diagnoses when using the query-based approach. When relying on test case based debugging, in contrast, more than one quarter of the users selected their solution from a list of more than one diagnosis. In other words,
at a certain point they stopped specifying further test cases and considered it more efficient to inspect the candidate list. We interpret this as a possible sign that test case based debugging was more tiring, and thus more demanding for the users than query answering.

\begin{figure}[ht]
	\centering
	\footnotesize
	\includegraphics[width=.8\linewidth]{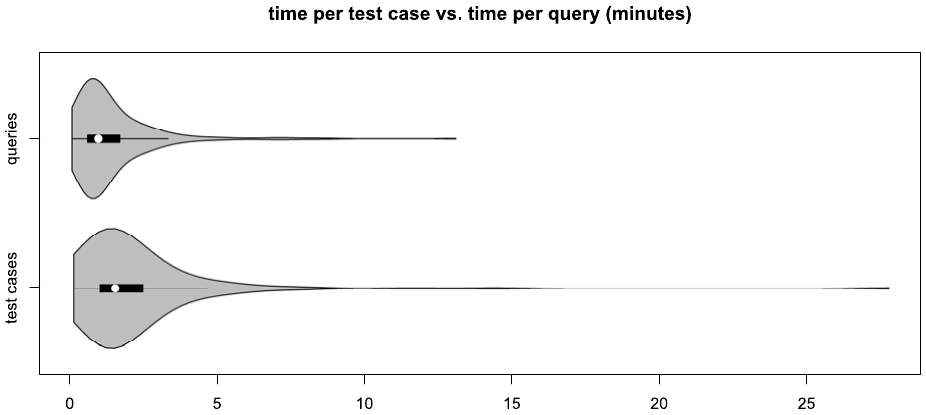}
	\caption{Violin plots showing the distribution of the time participants required to specify a test case
%	(test cases approach)
	vs.\ the time they required to answer a query
%	(queries approach)
	in \emph{Study~2}.}
	\label{fig:study2_vioplots_perQTtimes}
\end{figure}

\noindent\emph{Query answering is more efficient than test case specification:}
As both the query-based and the test case based approach result in the addition of a new test case per iteration\footnote{In the query-based scenario the test case is selected by the debugger and classified (as positive or negative) by the user, whereas in the test case based scenario the test case itself and its classification is chosen by the user.}, we compared the time users needed per answered query and per specified test case, respectively.
The result is very clear (cf.\ Figure~\ref{fig:study2_vioplots_perQTtimes}). The average test case specification time ($\approx$2:20 min) was almost 60\,\% (and statistically significantly\footnote{According to a Wilcoxon Rank-Sum Test with $\alpha=0.001$ (p-value = $8.96*10^{-13}$).})
%\footnote{According to %both a Wilcoxon Rank-Sum Test and
%a Student's t-Test with $\alpha=0.001$ (p-value = 0.0000071).})
%%TODO add test to verify normal distribution and equal variance
%\tododj{removed wilcoxon test. we only need one.}
higher than the average query answering time ($\approx$1:30 min).\footnote{Standard deviation: $\approx$1:30 (queries) and $\approx$2:50 (test cases).} This shows that it is more efficient to classify pre-selected axioms as (non-)entailments than to think about specific axioms \emph{and} classifying them. Overall, this result demonstrates the potential of query-based sequential diagnosis approaches to reduce debugging efforts.
%\tododj{I have rephrased things a bit and removed the many references. I guess they are all mentioned somewhere else already.}
%This shows that research on sequential diagnosis \cite{dekleer1987,Rodler2013,rodler-dx17,rodler17dx_activelearning,rodler2018ruleML,Shchekotykhin2012,Siddiqi2011,gonzalez2011spectrum}, i.e., on ways of and heuristics for selecting gainful queries to minimize the debugging effort, is indeed pivotal.

\noindent\emph{Query optimization pays off:} The average number of queries (11.6) that had to be answered until the true diagnosis was found by the users was lower than the average number of test cases (13.1) the users specified to isolate the true diagnosis.\footnote{Stadard deviation: 3.3 (queries) and 4.8 (test cases).}
This shows that automatic (and optimized\footnote{We used entropy-based query optimization as described in \cite{Shchekotykhin2012} in our study.}) test case selection tends to be more efficient than manual test case specification. In other words, the automated approach is better than users at selecting test cases that discriminate (well) between the candidates.

\subsubsection{Existence of Oracle Errors (RQ3.2)}
\label{sssec:existence_of_oracle_errors}

Both in \emph{Study 1} and \emph{Study 2}, we observed that it is not uncommon that participants make errors when specifying test cases and when answering the system's queries. While in either case the large majority of the inputs provided by the participants was correct, at least one mishap occurred to a considerable fraction of participants in both studies. Even in the main \emph{Study 2}, where the participants were instructed more intensively and where the participants had a better formal education on ontology engineering, about one quarter of the participants made at least one mistake. In the context of the study, mistakes were made equally for the test case specification and the query answering tasks.

Our observations therefore point to a largely open issue in algorithmic testing and debugging approaches, which are usually based on the assumption that there are no oracle errors. Only a few works exist in the literature, which specifically address the problem of wrong user inputs, e.g., in the context of spreadsheet testing \cite{Ruthruff:2005:ESF:1062455.1062523}, Spectrum-based Fault Localization procedures \cite{DBLP:conf/qrs/HoferW15}, or general software testing \cite{DebugOracle2016}.

Next, in Section \ref{sec:prediction-model}, we will take first steps to address this largely open research question in the context of query-based knowledge base debugging. Specifically, we will describe an initial prediction model that allows us to estimate the probability of oracle errors depending on the complexity of the queries asked to the user.

\section{Predicting Oracle Errors based on Query Complexity}
\label{sec:prediction-model}
When designing a query-based debugging method, different options are available with respect to what types of queries are asked to the users. A closer look at the wrong user inputs in \emph{Study 1} and \emph{Study 2} revealed that from the faulty test case specifications about two thirds had a non-trivial syntactic structure, involving, for example, complex class expressions with intersection, union, or complement operators, as defined in the OWL specification \cite{OWL2specification}. This supports the intuitive assumption that the syntactic complexity of the required inputs is correlated with the probability of a user error.

The goal of the work described in this section is to develop a first model that allows us to estimate the probability of user error for a given query.
%in a quantitative way.
The model can then be used by designers of interactive debugging systems, for example, in order to vary the complexity of the queries depending on the assumed expertise of the user. Alternatively, the model can be used to provide additional hints to the user in case of complex queries.

The proposed model was developed and evaluated with the help of two additional studies, which were performed in the context of \emph{Study 1} and \emph{Study 2}. The first of these studies, termed \emph{Study E1}, aimed at 
\begin{enumerate*}[label=\textit{(\roman*)}]
	\item verifying the conjecture that an axiom's syntactic complexity has indeed a \emph{significant} impact on how well it is understood and 
	\item collecting data as a basis for the design of the prediction model. The second study, termed \emph{Study E2}, was conducted to assess the utility of the model.
\end{enumerate*}
%\todoil{currently we use ``class'' (OWL) and ``concept'' (DL) interchangeably...maybe we should consistently use only one of these?!}
%\tododj{Yes, please harmonize this. Not sure what is better. We should check if we use the term concept in a more informal way somewhere else.}

\subsection{Collecting Data for the Prediction Model (Study E1)}
We designed a pen-and-paper study, where the task of the participants---the same ones as in \emph{Study 1}---was to determine the correct translation of axioms written in OWL (Manchester Syntax \cite{horridge2006manchester}) into natural language and vice versa. Each participant was provided with ten axioms that were randomly chosen from a larger pool of manually-prepared axioms. The axioms themselves, which again related to the university and IT domain, were designed to have different complexity levels. A simple axiom, for example, would be \stt{X~SubClassOf~Y}, where \stt{X} and \stt{Y} are class names from the respective domain.
More sophisticated axioms involved complex class expressions such as~\stt{not(X~and~Y)}~or~\stt{p some (X or Y)} which use, e.g., property restrictions and different logical operators.
An example of a more complex axiom would be
~\stt{UndergradStudent SubClassOf not (hasDegree some~Degree)}.
%\todoil{Attention: We use a different notation for OWL axioms in Tab. 1 and here...}
%\tododj{Please fix.}

For each given axiom, the participants were provided with three possible translations, where only one of them was correct. They then had to assign
confidence scores
to these answer options that express their degree of belief in the correctness of the respective answer.

To verify our hypothesis that syntactically more complex axioms are more difficult to comprehend, we proceeded as follows.
First, we gathered the confidence scores the participants gave to the correct answers for all the translation tasks. Next, we asked two experts to classify the syntax patterns that occurred in the exercises
as either particularly hard or particularly easy or neither.
We then compared the recorded confidence scores between the group of hard and the group of easy syntax patterns. The average score was 0.55 for the former and 0.95 for the latter group.
The statistical significance of this difference was revealed by a Wilcoxon Rank-Sum Test with level $\alpha = 0.01$ (p-value = 0.0015). That is, axioms of higher complexity indeed led to a lower success rate of the translation task.
Overall, this finding supports the relevance of a syntax-based prediction model.

To obtain further insights regarding which syntactic features cause difficulties for the users, we manually inspected all answers of the participants. As a result, we identified the following major factors that increase the complexity for the participants: (a) nesting of class expressions, (b) negation in general, and (c) negated expressions that are not represented in ``negation normal form'' (NNF), i.e., which include negated complex class expressions.

\subsection{Design of the Prediction Model}
Based on the lessons learned from the different studies
and on our researcher expertise,
we constructed a rule-based prediction model, which takes a query in OWL as an input and returns a score that expresses how likely it is that the query will be properly understood.
In other words, the model will tell us the likelihood of an oracle error for the given query.

The idea of the model is to recursively reduce a query to the axioms it consists of, and to then decompose these axioms to the class expressions they comprise. These expressions are in turn successively split into smaller sub-expressions, and so forth, until atomic classes are obtained. Based on the encountered syntactic structure,
the model uses respective weights
to compute the final query score when the recursion unwinds.
The weights are defined based on the observations of our study.

For instance, the model assigns \texttt{\small{X~SubClassOf~Y}} a score of $1$ (maximum ``easiness'') because such axioms were always correctly understood by the participants. In contrast, the score for \texttt{\small{X SubClassOf not (p some Z)}} would be 0.25 due to the involved negation and property restriction. Note that the axiom \texttt{\small{X SubClassOf p only (not Z)}} that expresses the same fact but is written differently in NNF would be indeed rated as being easier (score 0.29) by the model, which is in accordance with our observations.

To initially validate our model, we performed a correlation analysis based on \emph{Study E1}. The analysis revealed that the predictions for the exercises from \emph{Study E1}
are well in line with the success rates we had observed in the study (Pearson's $r=0.53$).
For the sake of brevity, we only sketched the main idea of the model here. The exact definition of the model can be found in \ref{apx:prediction_model}.
%\todoil{Add argumentation somewhere which addresses the possible objection that we could automatically translate the axioms in queries to natural language (there are such approaches) and ask the oracle simply the natural language sentence. The idea could be to argue that it is not only the correct translation that matters, but also the complexity of the question itself. E.g., one might be able to correctly translate an OWL axiom, but the axiom's difficulty might still prevent the oracle from answering correctly.}
%\todoil{State which exact OWL (sub-)language our model can currently handle? State that we currently support only terminological axioms (class expression axioms), no R-Box axioms (object/data property axioms)? }

\subsection{Evaluation of the Prediction Model (Study E2)}
\emph{Study E2}, which involved the participants of \emph{Study~2}, was a pen-and-paper exercise that we conducted to validate the predictive power of our model directly, i.e., through a query answering task.
In the study, each participant was provided with a natural language description of a university domain and 25 queries in OWL Manchester Syntax, each consisting of one axiom. The queries were randomly selected from a pool of
logical axioms $\tax_i$ involving $51$ syntactic patterns of different complexities, with scores predicted by our model ranging from
0.05 (hard) to 1 (easy).
For each query, the task was to decide if it is true or false in the given domain. The correct answers to all 25 questions were given in the natural language text, i.e., the participants did not have to make any assumptions to correctly answer the queries. The participants were again asked to provide, for each query, on a scale from 0 to 100, \begin{enumerate*}[label=\textit{(\roman*)}]
	\item a difficulty assessment and 
	\item their confidence in the given answer.
\end{enumerate*}

From the subjects' questionnaires, we extracted, grouped by syntactic pattern, (a)~the percentage of correct answers, (b)~the users' average confidence in their answer, and (c)~the average subjective difficulty. A comparison of each of these three response variables with the model predictions yielded quite decent correlation coefficients
of 0.36, 0.52, -0.70 for (a), (b) and (c), respectively.
Moreover, to assess the statistical significance of the model's predictive power, we ranked all queries according to their score as per our prediction model and performed a median split of the axioms into two groups, one including the easy and one the hard syntactic patterns.
An analysis of the response variables (a), (b) and (c) for these two groups revealed that there is a significant between-group difference (Wilcoxon Rank-Sum Test, p-values $<10^{-5}$, $<10^{-5}$ and $0.0197$)
which confirms the %expediency
predictive power of the proposed model.
As a result, axioms that were estimated to be hard according to the model
\begin{enumerate*}[label=\textit{(\roman*)}]
\item in fact led to a higher failure rate, 
\item were actually perceived to be harder, and 
\item resulted in a lower confidence of the users in their answers.
\end{enumerate*} The same relationship holds in the other direction.

As a side note, the prediction model, in case it did not exactly predict the observed success rate,
tended to underestimate the success probability. As a consequence, whenever the model predicted that a query is easy (i.e., had a score close to $1$), it actually proved to be \emph{very well}
understood by the users.
Hence, using methods in a query-based debugger that are able to generate
``easy questions'' with respect to such a prediction model is expected to be beneficial to avoid oracle errors. Examples of such methods can be found in \cite{rodler_jair-2017,rodler-dx17,DBLP:journals/corr/Rodler16a}.

\subsection{Discussion}
Overall, our results indicate that our model, although still preliminary, is able to assess the complexity of a given query with good reliability. Clearly, more research is required to further develop the model and to validate it for other problem settings. Nonetheless, we see the results as an important first step in the direction, which can be used when designing an interactive debugging environment.

Furthermore, the model can also be used for other purposes related to debugging, e.g., as an estimator of the prior fault information provided to a debugger. For instance, a higher fault probability could be assigned to axioms in the KB that are rated as hard by the prediction model. As pointed out and empirically proven by several works \cite{Rodler2013,Shchekotykhin2012,rodler2018ruleML}, reliable fault probabilities are a crucial ingredient to efficient fault localization but are often difficult to estimate.

\section{Research Limitations}
Our research does not come without limitations. First, the number of participants in the different studies, while being larger than in some previous studies on the topic, could be higher, and we plan to conduct additional experiments in the future with a larger set of participants. The participants of our studies were computer science students and all had a comparable background. We argue that this participant group is representative of at least a part of the population of real-world knowledge engineers, i.e., those that have a formal education in computer science.

%%%%% OLD text START %%%%%%%%%%%%%%%%%%%%
%The experiments conducted in \emph{Study 1} and \emph{Study 2} are each based on two specific knowledge bases (ontologies).
%While we thereby tried to make sure that the insights are not limited to one single domain, our experiments were based on ontologies with a comparable level of complexity. To what extent our insights generalize to much larger knowledge bases, therefore cannot be concluded with certainty from the made experiments.
%%%%% OLD text END %%%%%%%%%%%%%%%%%%%%
The experiments conducted in \emph{Study 1} and \emph{Study 2} are each based on two specific knowledge bases (ontologies).
While we thereby tried to make sure that the insights are not limited to one single domain, our experiments were based on ontologies with a comparable level of complexity. To what extent our insights generalize to much larger or more complex knowledge bases, can therefore not be concluded with certainty from the made experiments.
%However, based on our analyses regarding the time users required \emph{per test case} versus the time they needed \emph{per query} (cf.\ Section~\ref{sec:additional_obs_study2}), which was significantly (almost 40\,\%) less in the case of queries, and based on the comparable but by tendency smaller number of queries than test cases per debugging session, we believe that the higher efficiency of the query-based approach will in particular be present in case of harder debugging problems (involving a higher number of fault candidates and thus requiring even more user interactions).
%
%However, based on (i)~our analyses regarding the time users required \emph{per test case} and \emph{per query} (cf.\ Section~\ref{sec:additional_obs_study2}), which was significantly (almost 40\,\%) less in the case of queries, and based on (ii)~the comparable but by tendency smaller number of queries than test cases per debugging session, we believe that the higher efficiency of the query-based approach will be particularly present in case of harder debugging problems (involving more fault candidates and thus requiring more user interactions).
%
%However, the following aspects make us believe that the efficiency results carry over to harder debugging problems as well.
However, in the light of the following considerations it seems plausible
%TODO should we say it even more cautiously, e.g., ``...would be not very surprising...'' ??
to expect that the obtained results regarding debugging efficiency carry over to harder debugging problems as well.
First, we used ontologies, which are already highly expressive ($\mathcal{SROIQ}$) in Description Logic terms and hence simulate scenarios where users are confronted with very complex problems from the comprehension and reasoning point of view.
%
%do believe that the higher efficiency of the query-based approach will be particularly present in case of harder debugging problems (involving more fault candidates and thus requiring more user interactions).
%
Second, we observed that users required 
\begin{enumerate*}[label=\textit{(\roman*)}]
	\item significantly (almost 40\,\%) less time \emph{per query} than \emph{per test case}, and 
	\item a comparable but by tendency smaller number of queries than test cases per debugging session. 
\end{enumerate*}
This suggests growing (absolute) time savings of the query-based over the test case based approach when larger debugging problems involving more fault candidates and more user interactions are considered.
%%%%%%%%%%%%%%%%%%%%%%%%%%%%%%%%%
%Second, the significantly (almost 40\,\%) shorter time users required \emph{per query} than \emph{per test case},  and the comparable but by tendency smaller number of queries than test cases per debugging session, suggest at least the efficiency gains of the query-based over the test case based approach we observed when larger debugging problems involving more fault candidates and more user interactions are considered.
%%%%%%%%%%%%%%%%%%%%%%%%%%%%%%%%%
%Second, our findings with regard to (i)~the time users required \emph{per test case} and \emph{per query} (cf.\ Section~\ref{sec:additional_obs_study2}), which was significantly (almost 40\,\%) less in the case of queries, and (ii)~the comparable but by tendency smaller number of queries than test cases per debugging session, suggest at least the efficiency gain
%
%higher efficiency of the query-based approach will be particularly present in case of larger problems (involving more fault candidates and thus more user interactions).

The prediction model presented in Section \ref{sec:prediction-model} is still preliminary and must be seen more as a \emph{general} indicator than a precise, optimized predictor. In fact,
the scores that describe the complexity of an axiom are, for now, estimates that are based on a single study and on our own researcher expertise. However, our model evaluation clearly indicates that the rules, i.e., \emph{the way of} using the structure of an axiom for the estimation (e.g., deeper nesting of sub-clauses is harder), are plausible.

\section{Summary}
Tool support for debugging is not only relevant for traditional software systems, but also for knowledge-based systems. In the field of general software engineering, more and more research works are published which aim at better understanding the true value of such debugging tools for developers. In the field of knowledge-based systems, research on this topic is however still limited. With this work, we aim to contribute new insights regarding the usefulness of query-based knowledge base debugging in contrast to a more traditional test case based approach.

We conducted different user studies to address some of the open questions. The studies showed that users who were supported by any of the two forms of a model-based debugger were able to successfully locate a large fraction---in one study almost all---of the faults in the given knowledge bases. This emphasizes the usefulness of model-based knowledge debugging in general.
The query-based approach furthermore proved to be advantageous in terms of the efficiency and, thus, the required user effort in the debugging process. Users not only needed less time and fewer mouse clicks to locate the faults, the internal, optimizing query selection strategy also reduces the number of test cases that are needed to isolate the true cause of the observed problems.

Finally, the studies revealed certain other phenomena of knowledge base debugging processes. One main insight is that measuring the capability of a debugging method to properly rank the fault candidates should not be the only measure to compare different strategies. Another important aspect is that users sometimes provide wrong inputs to the debugging process. Future debuggers should therefore be able to take this aspect into account. In this work, we made a first step in this direction and proposed and evaluated a model that predicts the reliability of the user input for a query of a given complexity. Such predictions can, for example, be used in future systems to decide on which types of queries should be asked to the user in query-based approaches.

\section*{Acknowledgements}
This work was supported by the Carinthian Science Fund (KWF), contract KWF-3520/26767/38701.

%One observation of our studies was that users sometimes provide wrong inputs to the debugging process. In our work, we have designed an initial model to predict the likelihood of oracle errors based on the complexity of the queries. We provided initial evidence for the usefulness of such a model, which can be used to inform the design of future interactive debugging approaches.

%\tododj{Add a few sentence on future works / logical next steps? Only required if we have not done so throughout the text.}

%\section*{References}
%\bibliography{literature}

\begin{thebibliography}{10}
\expandafter\ifx\csname url\endcsname\relax
  \def\url#1{\texttt{#1}}\fi
\expandafter\ifx\csname urlprefix\endcsname\relax\def\urlprefix{URL }\fi
\expandafter\ifx\csname href\endcsname\relax
  \def\href#1#2{#2} \def\path#1{#1}\fi


\bibitem{rodler2019userstudy}
P.~Rodler, D.~Jannach, K.~Schekotihin, P.~Fleiss, {Are query-based ontology debuggers really helping knowledge engineers?}, Knowledge-Based Systems 179 (2019) 92--107.

\bibitem{ModernApproach}
S.~J. Russell, P.~Norvig, {Artificial Intelligence: A Modern Approach (Third
  Edition)}, Pearson, 2015.

\bibitem{DLHandbook}
F.~Baader, D.~Calvanese, D.~McGuinness, D.~Nardi, P.~Patel-Schneider (Eds.),
  {The Description Logic Handbook: Theory, Implementation, and Applications},
  Cambridge University Press, 2003.

\bibitem{pinedo2016scheduling}
M.~L. Pinedo, Scheduling: theory, algorithms, and systems, 5th Edition,
  Springer, 2016.

\bibitem{felfernig2014knowledge}
A.~Felfernig, L.~Hotz, C.~Bagley, J.~Tiihonen, Knowledge-based configuration:
  From research to business cases, Morgan Kaufmann, 2014.

\bibitem{Jannach2010}
D.~Jannach, M.~Zanker, A.~Felfernig, G.~Friedrich, Recommender Systems: An
  Introduction, Cambridge University Press, 2010.

\bibitem{Ceraso71}
J.~Ceraso, A.~Provitera, {Sources of error in syllogistic reasoning}, Cognitive
  Psychology 2~(4) (1971) 400--410.

\bibitem{Johnson1999}
P.~N. Johnson-Laird, {Deductive reasoning}, Annual review of psychology 50
  (1999) 109--135.

\bibitem{Rector2004}
A.~Rector, N.~Drummond, M.~Horridge, J.~Rogers, H.~Knublauch, R.~Stevens,
  H.~Wang, C.~Wroe, {OWL Pizzas: Practical Experience of Teaching OWL-DL:
  Common Errors \& Common Patterns}, in: 14th International Conference
  Engineering Knowledge in the Age of the Semantic Web (EKAW 2004), 2004, pp.
  63--81.

\bibitem{Roussey2009}
C.~Roussey, O.~Corcho, L.~M. Vilches-Bl\'{a}zquez, {A catalogue of OWL ontology
  antipatterns}, in: International Conference On Knowledge Capture (K-CAP
  2009), 2009, pp. 205--206.

\bibitem{Noy2006a}
N.~F. Noy, A.~Chugh, W.~Liu, M.~A. Musen, A framework for ontology evolution in
  collaborative environments, in: International Semantic Web Conference (ISWC
  2006), 2006, pp. 544--558.

\bibitem{ji2009radon}
Q.~Ji, P.~Haase, G.~Qi, P.~Hitzler, S.~Stadtm{\"u}ller, {RaDON—repair and
  diagnosis in ontology networks}, in: European Semantic Web Conference (ESWC
  2009), Springer, 2009, pp. 863--867.

\bibitem{meilicke2011}
C.~Meilicke, {Alignment Incoherence in Ontology Matching}, Ph.D. thesis,
  Universit{\"a}t Mannheim (2011).

\bibitem{Reiter87}
R.~Reiter, {A Theory of Diagnosis from First Principles}, Artificial
  Intelligence 32~(1) (1987) 57--95.

\bibitem{paper:felfernig:2004}
A.~Felfernig, G.~Friedrich, D.~Jannach, M.~Stumptner, {Consistency-based
  diagnosis of configuration knowledge bases}, Artificial Intelligence 152~(2)
  (2004) 213--234.

\bibitem{DBLP:conf/aadebug/MateisSWW00}
C.~Mateis, M.~Stumptner, D.~Wieland, F.~Wotawa, {Model-Based Debugging of Java
  Programs}, in: Fourth International Workshop on Automated Debugging (AADEBUG
  2000), 2000.

\bibitem{DBLP:conf/semweb/FriedrichS05}
G.~Friedrich, K.~M. Shchekotykhin, {A General Diagnosis Method for Ontologies},
  in: 4th International Semantic Web Conference (ISWC 2005), 2005, pp.
  232--246.

\bibitem{JannachSchmitz2014}
D.~Jannach, T.~Schmitz, Model-based diagnosis of spreadsheet programs: a
  constraint-based debugging approach, Automated Software Engineering 23~(1)
  (2014) 105--144.

\bibitem{Friedrich1999}
G.~Friedrich, M.~Stumptner, F.~Wotawa, {Model-based diagnosis of hardware
  designs}, Artificial Intelligence 111~(1-2) (1999) 3--39.

\bibitem{10.1007/BFb0019402}
L.~Console, G.~Friedrich, D.~T. Dupr{\'e}, Model-based diagnosis meets error
  diagnosis in logic programs, in: Automated and Algorithmic Debugging (AADEBUG
  1993), 1993, pp. 85--87.

\bibitem{Rodler2015phd}
P.~Rodler, {Interactive Debugging of Knowledge Bases}, Ph.D. thesis,
  Alpen-Adria Universit\"at Klagenfurt (2015).

\bibitem{dekleer1987}
J.~de~Kleer, B.~C. Williams, {Diagnosing multiple faults}, Artificial
  Intelligence 32~(1) (1987) 97--130.

\bibitem{DBLP:conf/ecai/ShchekotykhinFRF14}
K.~M. Shchekotykhin, G.~Friedrich, P.~Rodler, P.~Fleiss, Sequential diagnosis
  of high cardinality faults in knowledge-bases by direct diagnosis generation,
  in: {ECAI}, Vol. 263, 2014, pp. 813--818.

\bibitem{Rodler2011}
P.~Rodler, K.~Shchekotykhin, P.~Fleiss, G.~Friedrich, {Balancing Brave and
  Cautious Query Strategies in Ontology Debugging}, in: Joint Workshop on
  Knowledge Evolution and Ontology Dynamics (EvoDyn2011), 2011.

\bibitem{Shchekotykhin2012}
K.~Shchekotykhin, G.~Friedrich, P.~Fleiss, P.~Rodler, {Interactive Ontology
  Debugging: Two Query Strategies for Efficient Fault Localization}, Journal of
  Web Semantics 12-13 (2012) 88--103.

\bibitem{DBLP:conf/foiks/SchekotihinRS18}
K.~Schekotihin, P.~Rodler, W.~Schmid, Ontodebug: Interactive ontology debugging
  plug-in for prot{\'{e}}g{\'{e}}, in: 10th International Symposium on
  Foundations of Information and Knowledge Systems (FoIKS 2018), 2018, pp.
  340--359.

\bibitem{DBLP:conf/icbo/SchekotihinRSHT18a}
K.~Schekotihin, P.~Rodler, W.~Schmid, M.~Horridge, T.~Tudorache, A
  prot{\'{e}}g{\'{e}} plug-in for test-driven ontology development, in:
  Proceedings of the 9th International Conference on Biological Ontology
  {(ICBO} 2018), Corvallis, Oregon, USA, August 7-10, 2018., 2018.

\bibitem{noy2003protege}
N.~F. Noy, M.~Crub{\'e}zy, R.~W. Fergerson, H.~Knublauch, S.~W. Tu,
  J.~Vendetti, M.~A. Musen, Prot{\'e}g{\'e}-2000: an open-source
  ontology-development and knowledge-acquisition environment, in: AMIA 2003
  Open Source Expo, 2003, pp. 953--953.

\bibitem{Parnin:2011:ADT:2001420.2001445}
C.~Parnin, A.~Orso, Are automated debugging techniques actually helping
  programmers?, in: International Symposium on Software Testing and Analysis
  (ISSTA '11), 2011, pp. 199--209.

\bibitem{DBLP:conf/euromicro/RamlerWS12}
R.~Ramler, D.~Winkler, M.~Schmidt, Random test case generation and manual unit
  testing: Substitute or complement in retrofitting tests for legacy code?, in:
  38th Euromicro Conference on Software Engineering and Advanced Applications
  (SEAA 2012), 2012, pp. 286--293.

\bibitem{DBLP:conf/issta/StaatsHKR12}
M.~Staats, S.~Hong, M.~Kim, G.~Rothermel, Understanding user understanding:
  Determining correctness of generated program invariants, in: International
  Symposium on Software Testing and Analysis (ISSTA 2012), 2012, pp. 188--198.

\bibitem{DBLP:conf/issta/FraserSMAP13}
G.~Fraser, M.~Staats, P.~McMinn, A.~Arcuri, F.~Padberg, Does automated
  white-box test generation really help software testers?, in: International
  Symposium on Software Testing and Analysis (ISSTA 2013), 2013, pp. 291--301.

\bibitem{Kalyanpur.Just.ISWC07}
A.~Kalyanpur, B.~Parsia, M.~Horridge, E.~Sirin, {Finding all Justifications of
  OWL DL Entailments}, in: 6th International Semantic Web Conference (ISWC
  2007), 2007, pp. 267--280.

\bibitem{Horridge2008}
M.~Horridge, B.~Parsia, U.~Sattler, {Laconic and Precise Justifications in
  OWL}, in: 7th International Semantic Web Conference (ISWC 2008), 2008, pp.
  323--338.

\bibitem{Junker04}
U.~Junker, {QUICKXPLAIN:} preferred explanations and relaxations for
  over-constrained problems, in: {AAAI}, {AAAI} Press / The {MIT} Press, 2004,
  pp. 167--172.

\bibitem{DBLP:conf/wlp/GebserPSTW07}
M.~Gebser, J.~P{\"u}hrer, T.~Schaub, H.~Tompits, S.~Woltran, spock: A debugging
  support tool for logic programs under the answer-set semantics, in:
  Applications of Declarative Programming and Knowledge Management, 2009, pp.
  247--252.

\bibitem{DBLP:journals/tplp/OetschPT10}
J.~Oetsch, J.~P{\"{u}}hrer, H.~Tompits, Catching the ouroboros: On debugging
  non-ground answer-set programs, {Theory and Practice of Logic Programming}
  10~(4-6) (2010) 513--529.

\bibitem{ParsiaSK05}
B.~Parsia, E.~Sirin, A.~Kalyanpur, Debugging {OWL} ontologies, in: Proceedings
  of the 14th international conference on World Wide Web, 2005, pp. 633--640.

\bibitem{SchlobachHCH07}
S.~Schlobach, Z.~Huang, R.~Cornet, F.~van Harmelen, Debugging incoherent
  terminologies, Journal of Automated Reasoning 39~(3) (2007) 317--349.

\bibitem{SchlobachC03}
S.~Schlobach, R.~Cornet, Non-standard reasoning services for the debugging of
  description logic terminologies, in: Proceedings of the 18th International
  Joint Conference on Artificial Intelligence, 2003, pp. 355--362.

\bibitem{Kalyanpur2006a}
A.~Kalyanpur, {Debugging and Repair of OWL Ontologies}, Ph.D. thesis,
  University of Maryland, College Park (2006).

\bibitem{BaaderP07}
F.~Baader, R.~Pe{\~n}aloza, Axiom pinpointing in general tableaux, in:
  Proceedings of the 16th International Conference on Automated Reasoning with
  Analytic Tableaux and Related Methods, 2007, pp. 11--27.

\bibitem{BaaderP10}
F.~Baader, R.~Pe{\~n}aloza, Automata-based axiom pinpointing, J. Autom.
  Reasoning 45~(2) (2010) 91--129.

\bibitem{ChengQ11}
X.~Cheng, G.~Qi, An algorithm for axiom pinpointing in {EL+} and its
  incremental variant, in: 20th ACM Conference on Information and Knowledge
  Management (CIKM 2011), 2011, pp. 2433--2436.

\bibitem{DBLP:conf/sum/OzakiP18}
A.~Ozaki, R.~Pe{\~{n}}aloza, Consequence-based axiom pinpointing, in: 12th
  International Conference on Scalable Uncertainty Management (SUM 2018), 2018,
  pp. 181--195.

\bibitem{DBLP:conf/cade/KazakovS18}
Y.~Kazakov, P.~Skocovsk{\'{y}}, Enumerating justifications using resolution,
  in: 9th International Joint Conference on Automated Reasoning (IJCAR 2018),
  2018, pp. 609--626.

\bibitem{DBLP:journals/ai/PenalozaS17}
R.~Pe{\~{n}}aloza, B.~Sertkaya, Understanding the complexity of axiom
  pinpointing in lightweight description logics, Artificial Intelligence 250
  (2017) 80--104.

\bibitem{KalyanpurPHS07}
A.~Kalyanpur, B.~Parsia, M.~Horridge, E.~Sirin, Finding all justifications of
  {OWL} {DL} entailments, in: Proceedings of the 6th International Semantic Web
  Conference, 2007, pp. 267--280.

\bibitem{DBLP:conf/aaai/Shchekotykhin15}
K.~M. Shchekotykhin, Interactive query-based debugging of {ASP} programs, in:
  29th AAAI Conference on Artificial Intelligence (AAAI 2015), 2015, pp.
  1597--1603.

\bibitem{DBLP:journals/ws/ShchekotykhinFFR12}
K.~M. Shchekotykhin, G.~Friedrich, P.~Fleiss, P.~Rodler, Interactive ontology
  debugging: Two query strategies for efficient fault localization, Journal of
  Web Semantics 12 (2012) 88--103.

\bibitem{DBLP:conf/kr/GrauJKZ12}
B.~C. Grau, E.~Jim{\'{e}}nez{-}Ruiz, E.~Kharlamov, D.~Zheleznyakov, Ontology
  evolution under semantic constraints, in: {KR}, {AAAI} Press, 2012.

\bibitem{DBLP:conf/tableaux/FurbachS13}
U.~Furbach, C.~Schon, Semantically guided evolution of aboxes, in: {TABLEAUX},
  Vol. 8123 of Lecture Notes in Computer Science, Springer, 2013, pp. 134--148.

\bibitem{rodler_jair-2017}
P.~Rodler, W.~Schmid, K.~Schekotihin, A generally applicable, highly scalable
  measurement computation and optimization approach to sequential model-based
  diagnosis, CoRR abs/1711.05508.

\bibitem{rodler17dx_activelearning}
P.~Rodler, On active learning strategies for sequential diagnosis, in: 28th
  International Workshop on Principles of Diagnosis (DX'17), Vol.~4, 2018, pp.
  264--283.

\bibitem{WangHRDS05}
H.~Wang, M.~Horridge, A.~L. Rector, N.~Drummond, J.~Seidenberg, Debugging
  {OWL-DL} ontologies: {A} heuristic approach, in: Proceedings of the 4th
  International Semantic Web Conference, 2005, pp. 745--757.

\bibitem{DBLP:conf/f-egc/RousseyCSSB12}
C.~Roussey, {\'{O}}.~Corcho, O.~Sv{\'{a}}b{-}Zamazal, F.~Scharffe, S.~Bernard,
  Antipattern detection in web ontologies: an experiment using {SPARQL}
  queries, in: {The Second International Workshop on Debugging Ontologies and
  Ontology Mappings (WoDOOM)}, 2012, pp. 45--56.

\bibitem{DBLP:journals/jamia/RectorBS11}
A.~L. Rector, S.~Brandt, T.~Schneider, Getting the foot out of the pelvis:
  modeling problems affecting use of {SNOMED} {CT} hierarchies in practical
  applications, {Journal of the American Medical Informatics Association}
  18~(4) (2011) 432--440.

\bibitem{DBLP:conf/cp/NethercoteSBBDT07}
N.~Nethercote, P.~J. Stuckey, R.~Becket, S.~Brand, G.~J. Duck, G.~Tack,
  Minizinc: Towards a standard {CP} modelling language, in: {Principles and
  Practice of Constraint Programming (CP 2007)}, 2007, pp. 529--543.

\bibitem{Musen2015}
M.~A. Musen, {The Prot{\'{e}}g{\'{e}} Project: A Look Back and a Look Forward},
  {AI} Matters 1~(4) (2015) 4--12.

\bibitem{DBLP:conf/lpnmr/FebbraroRR11}
O.~Febbraro, K.~Reale, F.~Ricca, Aspide: Integrated development environment for
  answer set programming, in: 11th International Conference on Logic
  Programming and Nonmonotonic Reasoning (LPNMR 2011), 2011, pp. 317--330.

\bibitem{DBLP:journals/tplp/WielemakerSTL12}
J.~Wielemaker, T.~Schrijvers, M.~Triska, T.~Lager, {SWI-Prolog}, {Theory and
  Practice of Logic Programming} 12~(1-2) (2012) 67--96.

\bibitem{dodaro2015interactive}
C.~Dodaro, P.~Gasteiger, B.~Musitsch, F.~Ricca, K.~Shchekotykhin, {Interactive
  debugging of non-ground ASP programs}, in: International Conference on Logic
  Programming and Nonmonotonic Reasoning (LPNR 2015), 2015, pp. 279--293.

\bibitem{DBLP:conf/cpaior/LeoT17}
K.~Leo, G.~Tack, Debugging unsatisfiable constraint models, in: Integration of
  AI and OR Techniques in Constraint Programming (CPAIOR 2017), 2017, pp.
  77--93.

\bibitem{DBLP:journals/ws/KalyanpurPSGH06}
A.~Kalyanpur, B.~Parsia, E.~Sirin, B.~C. Grau, J.~A. Hendler, {Swoop: {A} Web
  Ontology Editing Browser}, Journal of Web Semantics 4~(2) (2006) 144--153.

\bibitem{Horridge2011b}
M.~Horridge, S.~Bail, B.~Parsia, U.~Sattler, {The cognitive complexity of OWL
  justifications}, in: Proceedings of the 10th International Semantic Web
  Conference, 2011, pp. 241--256.

\bibitem{DBLP:conf/ekaw/Svab-ZamazalS08}
O.~{\v{S}}v{\'a}b-Zamazal, V.~Sv{\'a}tek, Analysing ontological structures
  through name pattern tracking, in: Knowledge Engineering: Practice and
  Patterns (EKAW 2008), 2008, pp. 213--228.

\bibitem{corcho2009pattern}
O.~Corcho, C.~Roussey, L.~M.~V. Bl{\'a}zquez, I.~P{\'e}rez, {Pattern-based OWL
  Ontology Debugging Guidelines}, in: Workshop on Ontology Patterns, 2009,
  p.~68.

\bibitem{SchekotihinSchmitzEtAl2016}
K.~Schekotihin, T.~Schmitz, D.~Jannach, Efficient sequential model-based
  fault-localization with partial diagnoses, in: International Joint Conference
  on Artificial Intelligence (IJCAI 2016), 2016, pp. 1251--1258.

\bibitem{rodler17dx_reducing}
P.~Rodler, K.~Schekotihin, Reducing model-based diagnosis to knowledge base
  debugging, in: 28th International Workshop on Principles of Diagnosis
  (DX'17), Vol.~4, 2018, pp. 284--296.

\bibitem{OWL2specification}
{OWL 2 Web Ontology Language. Structural Specification and Functional-Style
  Syntax (Second Edition), Online at: http://www.w3.org/TR/owl2-syntax/}.

\bibitem{qi2007measuring}
G.~Qi, A.~Hunter, Measuring incoherence in description logic-based ontologies,
  in: 6th International Semantic Web Conference (ISWC 2007), 2007, pp.
  381--394.

\bibitem{rodler2018socs}
P.~Rodler, M.~Herold, {StaticHS}: {A} variant of {Reiter}'s hitting set tree
  for efficient sequential diagnosis, in: 11th International Symposium on
  Combinatorial Search ({SOCS} 2018), 2018, pp. 72--80.

\bibitem{RodlerH18_dx}
P.~Rodler, M.~Herold, Reducing sequential diagnosis costs by modifying reiter's
  hitting set tree, in: 29th International Workshop on Principles of Diagnosis
  (DX'18) co-located with 10th {IFAC} Symposium on Fault Detection, Supervision
  and Safety for Technical Processes {(SAFEPROCESS}'18), 2018.

\bibitem{darwiche2001decomposable}
A.~Darwiche, Decomposable negation normal form, Journal of the ACM 48~(4)
  (2001) 608--647.

\bibitem{jiang2003computation}
Y.-F. Jiang, L.~Lin, The computation of hitting sets with boolean formulas,
  Chinese Journal of Computers 26~(8) (2003) 919--924.

\bibitem{torasso2006}
P.~Torasso, G.~Torta, {Model-based diagnosis through OBDD compilation: A
  complexity analysis}, in: Reasoning, Action and Interaction in AI Theories
  and Systems, 2006, pp. 287--305.

\bibitem{metodi2014}
A.~Metodi, R.~Stern, M.~Kalech, M.~Codish, A novel sat-based approach to model
  based diagnosis, Journal of Artificial Intelligence Research 51 (2014)
  377--411.

\bibitem{fikes2004owl}
R.~Fikes, P.~Hayes, I.~Horrocks, Owl-ql—a language for deductive query
  answering on the semantic web, Web semantics: Science, services and agents on
  the World Wide Web 2~(1) (2004) 19--29.

\bibitem{DBLP:journals/corr/Rodler16a}
P.~Rodler, Towards better response times and higher-quality queries in
  interactive knowledge base debugging, Tech. rep., Alpen-Adria Universit\"at
  Klagenfurt, http://arxiv.org/pdf/1609.02584v2.pdf (2016).

\bibitem{rodler_singleton-2019}
P.~Rodler, M.~Eichholzer, A new expert questioning approach to more efficient
  fault localization in ontologies, CoRR abs/1904.00317.

\bibitem{Rodler2013}
P.~Rodler, K.~Shchekotykhin, P.~Fleiss, G.~Friedrich, {RIO: Minimizing User
  Interaction in Ontology Debugging}, in: Web Reasoning and Rule Systems (RR
  2013), 2013, pp. 153--167.

\bibitem{rodler2018ruleML}
P.~Rodler, W.~Schmid, On the impact and proper use of heuristics in test-driven
  ontology debugging, in: Rules and Reasoning - Second International Joint
  Conference (RuleML+RR 2018), 2018, pp. 164--184.

\bibitem{RodlerS18_dx}
P.~Rodler, W.~Schmid, Comparing the performance of traditional and novel
  heuristics for sequential diagnosis, in: 29th International Workshop on
  Principles of Diagnosis (DX'18) co-located with 10th {IFAC} Symposium on
  Fault Detection, Supervision and Safety for Technical Processes
  {(SAFEPROCESS}'18), 2018.

\bibitem{settles2012}
B.~Settles, {Active Learning}, Morgan and Claypool Publishers, 2012.

\bibitem{rodler-dx17}
P.~Rodler, W.~Schmid, K.~Schekotihin, Inexpensive cost-optimized measurement
  proposal for sequential model-based diagnosis, in: 28th International
  Workshop on Principles of Diagnosis (DX'17), Vol.~4, 2018, pp. 200--218.

\bibitem{Shearer2008}
R.~Shearer, B.~Motik, I.~Horrocks, Hermit: {A} highly-efficient {OWL} reasoner,
  in: 5th Workshop on OWL: Experiences and Directions, Vol. 432 of {CEUR}
  Workshop Proceedings, 2008.

\bibitem{sirin2007pellet}
E.~Sirin, B.~Parsia, B.~C. Grau, A.~Kalyanpur, Y.~Katz, {Pellet: A practical
  OWL-DL reasoner}, Journal of Web Semantics 5~(2) (2007) 51--53.

\bibitem{horrocks2006even}
I.~Horrocks, O.~Kutz, U.~Sattler, {The Even More Irresistible SROIQ}, in: 10th
  International Conference on Principles of Knowledge Representation and
  Reasoning, 2006, pp. 57--67.

\bibitem{owl1.1_spec}
P.~F. Patel-Schneider, I.~Horrocks, B.~Motik, {OWL 1.1 Web Ontology Language:
  Structural Specification and Functional-Style Syntax}, online,
  \url{https://www.w3.org/Submission/owl11-overview/} (December 2006).

\bibitem{horridge2006manchester}
M.~Horridge, N.~Drummond, J.~Goodwin, A.~L. Rector, R.~Stevens, H.~Wang, {The
  Manchester OWL Syntax}, in: {OWL Experiences and Directions Workshop}, CEUR
  Workshop Proceedings, 2006.

\bibitem{bhattacharyya1977}
G.~Bhattacharyya, R.~Johnson, Statistical Concepts and Methods, John Wiley \&
  Sons Inc, 1977.

\bibitem{Ruthruff:2005:ESF:1062455.1062523}
J.~R. Ruthruff, M.~Burnett, G.~Rothermel, An empirical study of fault
  localization for end-user programmers, in: 27th International Conference on
  Software Engineering (ICSE '05), 2005, pp. 352--361.

\bibitem{DBLP:conf/qrs/HoferW15}
B.~Hofer, F.~Wotawa, Fault localization in the light of faulty user input, in:
  2015 {IEEE} International Conference on Software Quality, Reliability and
  Security ({QRS} 2015), 2015, pp. 282--291.

\bibitem{DebugOracle2016}
X.~Guo, M.~Zhou, X.~Song, M.~Gu, J.~Sun, First, debug the test oracle, IEEE
  Transactions on Software Engineering 41~(10) (2015) 986--1000.

\end{thebibliography}

\newpage
\begin{appendix}

	\section{Formal Characterization of the Complexity Prediction Model}
	\label{apx:prediction_model}
	The suggested prediction model for query complexity is a function $M$ that maps a query $Q$---consisting of a set of OWL\footnote{Whenever we write OWL in this section, we mean the OWL~2 Web Ontology Language, as specified in \cite{OWL2specification}.}
	%(terminological)
	%(class expression)
	axioms\footnote{Currently, the model supports only class expression axioms. It can however be extended to cover object property, data property and assertion axioms as well.}
	%in future work.},
	---to a real-valued score in $(0,1]$ where $1$ means maximally easy and $0$ maximally hard, respectively. Intuitively, $M(Q)$ can be interpreted as an estimate of the query's probability to be
	%answered correctly
	comprehended properly by a user.
	%\tododj{Maybe okay, but I am not sure that these really are/can be interpreted as probabilities.}
	%
	%The input to the prediction model for query complexity is a query, i.e.\ a set of OWL axioms, and its output is a real-valued score in $[0,1]$ assessing the complexity of the input query where $1$ and $0$ mean maximally easy vs.\ maximally hard.
	
	%The principle behind the model is the following. First, we assume that different expressions (logical operators, quantifiers, etc.) appearing in axioms have different complexities. Second, we use empirical weights to describe these complexities. Third, we rely on a set of manually defined recursive rules that incorporate the said weights to derive the query complexity via a stepwise decomposition of the entire query to the smallest expressions it contains.
	The assumption behind the model is that different expressions (logical operators, quantifiers, etc.) appearing in axioms have different complexities. To describe these complexities, we use a set of weights that are chosen empirically and based on our expertise. These weights are incorporated into a set of manually defined recursive rules. We use these rules to derive the complexity of a given query by decomposing it stepwise to its smallest components.
	%via a stepwise decomposition of the query to its smallest components. %expressions. it contains.
	% for expressions consisting of subexpressions.
	%The core of the function $M$ is constituted by a set of pre-defined rules that
	
	Computationally, the underlying idea of the model is to first extract from the query the axioms it consists of. Each of these axioms is then reduced to the class expressions it comprises. The class expressions are then recursively split into smaller sub-expressions until atomic classes are obtained. Based on the structure of the axiom found by this recursive reduction, the model uses the specified
	%uses respective
	%\tododj{What is meant by ``respective''? Where do they come from?}
	weights to compute the
	complexity
	%easiness (probability of being answered correctly)
	of the axiom. Finally, the complexities of all query axioms are combined to compute the final query score.
	
	%We explicate these computations in detail by means of a top-down approach.
	%The overall top-down computation approach can be summarized as follows.
	%That is, we first show how axiom complexities are used to determine the overall query complexity, then show how axiom complexities are derived based on the class expressions the axioms comprise, and finally show how the complexities of the class expressions are calculated.
	%That is, we first show
	In the following, we will describe in more detail
	\begin{enumerate*}[label=\textbf{(\Roman*)}]
		\item \label{enum:apx:query_complexity} how axiom complexities are used to determine the overall query complexity,
		\item \label{enum:apx:axiom_complexity} how axiom complexities are derived based on the class expressions occurring in them,
		%the axioms comprise,
		and
		\item \label{enum:apx:class_complexity} how complexities of class expressions are calculated.
	\end{enumerate*}
	% show  then show , and finally show
	% final  query score when the recursion unwinds.
	
	%Specifically,
	%To realize the computations, t
	The function $M$ makes use of two additional functions. The function $M_{ax}$ computes for a given OWL axiom its estimated probability in $(0,1]$ of being
	%answered
	understood
	correctly; $M_{ce}$ computes for a given OWL class expression its complexity in terms of a real number in $[1,\infty)$.
	
	\ref{enum:apx:query_complexity} Overall query complexity: Let $Q = \{\tax_1,\dots,\tax_k\}$ be a query consisting of the OWL axioms $\tax_i$, $i\in\{1,\dots,k\}$. Then, we define
	$$M(Q) := \prod_{i=1}^{k} M_{ax}(\tax_i)$$
	That is, the probability of $Q$ being answered correctly is equal to the probability of all axioms in $Q$ being answered correctly (assuming independence between the axioms).
	%More formally, the complexity function can be described as follows.
	
	% we use DL notation here for brevity, see DL handbook
	\ref{enum:apx:axiom_complexity} Axiom complexity: Let $\tax$ be an OWL (class expression) axiom.
	%Then either $\tax: X \sqsubseteq Y$ or $\tax: X \equiv Y$ for class expressions $X,Y$.
	An axiom $\tax$ has one of the following forms \cite{OWL2specification} for some integer $s \geq 2$ and arbitrary OWL class expressions $X_1,\dots,X_s$:
	%\tododj{Is the term ``shape'' defined? Or should we just say: one of the following forms.}
	\begin{itemize}[noitemsep]
		\item $X_1 \;\texttt{\small SubClassOf}\; X_2$
		\item $\texttt{\small EquivalentClasses}\; X_1 \dots X_s$
		\item $\texttt{\small DisjointClasses}\; X_1 \dots X_s$
		\item $\texttt{\small DisjointUnion}\; X_1 \dots X_s$
	\end{itemize}
	% or  or  or
	%for class expressions $X_1,\dots,X_s$.
	%\tododj{Where do the Xi to Xs come from, what is s?}
	We denote by $\mathit{CE}(\tax)$ the set of all class expressions
	%$X_i$
	occurring in $\tax$ and specify
	$$M_{ax}(\tax) := \prod_{X_i \in \mathit{CE}(\tax)} \frac{1}{M_{ce}(X_i)}$$
	That is, the probability of understanding the entire axiom is equal to the probability of properly comprehending
	%\tododj{of *properly* comprehending?}
	%both the left and right parts, i.e.\ class expressions, of
	all class expressions occurring in
	the axiom (assuming independence between the user's understanding of the individual expressions). The estimated probability of comprehending a class expression is inversely proportional to the complexity of the expression, as assessed by $M_{ce}$.
	
	\ref{enum:apx:class_complexity} Class expression complexity:\footnote{For brevity of notation we use Description Logic Syntax in the following description wherever possible. E.g., ``$\sqcap$'', ``$\sqcup$'', ``$\lnot$'' stand for the OWL Manchester Syntax keywords \texttt{and}, \texttt{or} and \texttt{not}, respectively. For details see \cite[Fig.~3]{horridge2006manchester}.} We define $M_{ce}$ recursively as follows.
	Let $X_1,X_2,X_3,X_4,X_5,X_6$ be (complex or atomic) OWL class expressions, $A$ an atomic OWL class, and $C_1,C_2$ complex OWL class expressions.
	With an atomic OWL class we associate a named class, $\top$, $\bot$, or an enumeration of individuals\footnote{OWL keyword \texttt{\scriptsize ObjectOneOf} \cite{OWL2specification}.}.
	%Note that OWL classes characterized as enumerations of individuals\footnote{OWL keyword \texttt{\small ObjectOneOf} \cite{OWL2specification}} are here counted among atomic classes.
	%\tododj{Not sure what the last sentence means}
	Further, let $r_o$ be an OWL object property, $r_d$ an OWL data property, $r$ an OWL (data or object) property, $R$ a data range, $\mathbf{Q} \in \{\forall, \exists\}$, $\mathbf{N} \in \{ =, \leq, \geq\}$,
	%$\mathbf{V} \in \{\texttt{\small hasValue}\}$
	as well as $m$ a non-negative integer, $v$ an individual and $l$ a literal.
	Then:
	%\footnote{For brevity of notation, we use Description logic syntax here. That is, e.g., ``$\sqsubseteq$'' stands for the OWL Manchester Syntax keyword \texttt{SubClassOf} and ``$\equiv$'' for \texttt{EquivalentTo}. Further, ``$\sqcap$'', ``$\sqcup$'', ``$\lnot$'' represent \texttt{and}, \texttt{or} and \texttt{not}, respectively. For details, see \cite[Fig.~3]{horridge2006manchester}.}
	
	\begin{align*}
	M_{ce}(A \sqcap C_1) =\,& M_{ce}(C_1 \sqcap A) = M_{ce}(A) \cdot (1+M_{ce}(C_1)) \\
	\,&\mbox{ if } C_1=X_3 \sqcup X_4 \\
	M_{ce}(C_1 \sqcap C_2) =\,& (1+M_{ce}(C_1)) \cdot (1+M_{ce}(C_2)) \\
	\,&\mbox{ if } C_1=X_3 \sqcup X_4, C_2=X_5 \sqcup X_6 \\
	M_{ce}(X_1 \sqcap X_2) =\,& M_{ce}(X_1) \cdot M_{ce}(X_2) \\
	M_{ce}(A \sqcup C_1) =\,& M_{ce}(C_1 \sqcup A) = M_{ce}(A) \cdot (1+M_{ce}(C_1)) \\
	\,&\mbox{ if }  C_1=X_3 \sqcap X_4 \\
	M_{ce}(C_1 \sqcup C_2) =\,& (1+M_{ce}(C_1)) \cdot (1+M_{ce}(C_2)) \\
	\,&\mbox{ if } C_1=X_3 \sqcap X_4, C_2=X_5 \sqcap X_6 \\
	M_{ce}(X_1 \sqcup X_2) =\,& M_{ce}(X_1) \cdot M_{ce}(X_2) \\
	M_{ce}(\mathbf{Q}\, r_o\, A) =\,& M_{ce}(\mathbf{N}\, m\, r_o\, A) = 1 + M_{ce}(A) \\
	M_{ce}(\mathbf{Q}\, r_o\, C_1) =\,& M_{ce}(\mathbf{N}\, m\, r_o\, C_1) = 2 + M_{ce}(C_1) \\
	%%%%%%%%%%%%%%%%%%%
	M_{ce}(\mathbf{Q}\, r_d\, R) =\,& M_{ce}(\mathbf{N}\, m\,r_d\, R) =  M_{ce}(\mathbf{N}\, m\,r) = 2 \\
	M_{ce}(\texttt{\footnotesize ObjectHasValue}\, r\, v) =\,& 2 \\
	M_{ce}(\texttt{\footnotesize ObjectHasSelf}\, r) =\,&2 \\ M_{ce}(\texttt{\footnotesize DataHasValue}\, r\, l) =\,&2 \\
	%%%%%%%%%%%%%%%%%%%
	M_{ce}(A) =\,& 1 \\
	M_{ce}(\lnot A) =\,& 1.25 \\
	%M_{ce}(\lnot A) \,&=\,& M_{ce}(A) \\
	M_{ce}(\lnot C_1) =\,& 2 \cdot M_{ce}(C_1)
	\end{align*}
	Importantly, each class expression $ce$ is evaluated from top to bottom, i.e., the first of the above equations that is applicable is used to assess $ce$.
	%\tododj{Not sure if this can be easily understood. Also, I think that this last mentioned aspect (we have a set of pre-defined ruels) is actually one of the most important ideas, which have never been mentioned above. I think one can start the appendix somehow as follows. 1) we assume that different operations/expressions have different complexities 2) we use empirical weights that describe these complexities 3) we use a set of manually defined rules to derive complexities for expressions consisting of subexpressions. }
	% dealing with number restrictions, data properties, domain/range axioms (can be expressed equivalently in terms of the axiom structure given above)
	% deal with equivClasses, disjClasses and disjUnion
	%function $D: \mathit{CE} \to  \mathbb{R}^{+}$ which quantifies the complexity of (arbitrary) class expressions (set $\mathit{CE}$).
	%
	%The underlying idea is that the probability of understanding the entire axiom is equal to the probability of comprehending both the left and right parts of the axiom (assuming independence between the parts).
	
\end{appendix}

\end{document}